%% file: main.tex
\documentclass[letterpaper, 10 pt, conference]{cls/ieeeconf}
\IEEEoverridecommandlockouts
\usepackage[table,usenames,dvipsnames]{xcolor}      % color
\usepackage[noadjust]{cite}
\usepackage{amsmath,amssymb,amsfonts,amsthm,dsfont,mathtools}
\usepackage{nicematrix}
\usepackage{comment}

\usepackage{adjustbox}
\usepackage{graphicx,tabularx,adjustbox,color}
\usepackage{multirow}
\usepackage[font=footnotesize]{caption}
\usepackage[font=footnotesize]{subcaption}

\usepackage{algorithmic}
\usepackage[ruled,vlined]{algorithm2e}
\usepackage{stackengine}

\allowdisplaybreaks
\renewcommand{\baselinestretch}{.985}
\usepackage[breaklinks=true, colorlinks, bookmarks=true, citecolor=Black, urlcolor=Violet,linkcolor=Black]{hyperref}

\DeclareMathOperator*{\argmin}{arg\,min}

\newtheorem{theorem}{Theorem}[section]
\newtheorem{proposition}[theorem]{Proposition}

\theoremstyle{remark}

\theoremstyle{definition}
\newtheorem{definition}[theorem]{Definition}

\newtheorem*{problem*}{Problem}
\newtheorem{method}[]{Procedure}
\theoremstyle{definition}

\newcommand{\TODO}[1]{{\color{red}#1}}

\newcommand{\NA}[1]{$\spadesuit$\footnote{\TODO{Nikolay: #1}}}

\input{cls/sym.tex}

\title{\LARGE \bf Safe Stabilizing Control for Polygonal Robots in Dynamic Elliptical Environments} 

\author{Kehan Long \qquad Khoa Tran \qquad Melvin Leok \qquad Nikolay Atanasov
\thanks{The authors are with the Contextual Robotics Institute, University of California San Diego, La Jolla, CA 92093, USA (e-mails: {\tt\small \{k3long,\allowbreak k2tran,\allowbreak mleok,\allowbreak natanasov\}@ucsd.edu}).}%
}

\begin{document}
\maketitle
\begin{abstract}
This paper addresses the challenge of safe navigation for rigid-body mobile robots in dynamic environments. We introduce an analytic approach to compute the distance between a polygon and an ellipse, and employ it to construct a control barrier function (CBF) for safe control synthesis. Existing CBF design methods for mobile robot obstacle avoidance usually assume point or circular robots, preventing their applicability to more realistic robot body geometries. Our work enables CBF designs that capture complex robot and obstacle shapes. We demonstrate the effectiveness of our approach in simulations highlighting real-time obstacle avoidance in constrained and dynamic environments for mobile robots and 2-D robot arms. 
%bridges this gap by leveraging the proposed analytic distance computation to handle more complex robot and obstacle shapes. We demonstrate the effectiveness of our approach through numerical simulations, highlighting real-time obstacle avoidance capabilities in tight and dynamic environments for both ground robots and multi-link robot arms. 
\end{abstract}

\input{tex/Intro.tex}

\input{tex/ProblemFormulation.tex}

\input{tex/Prelim.tex}
\input{tex/Polygon_Ellipse_Env}

\input{tex/MultipleObstacle.tex}

%\input{tex/Quadrotor}
\input{tex/Evaluation.tex}
\input{tex/Conclusion.tex}

%==================================================================%
% References

\bibliography{ref}
\bibliographystyle{ieeetr}

%\appendices
% \setcounter{section}{0}
% \renewcommand{\thesection}{Appendix \Alph{section}}
%%\input{tex/appendix_1}
%\input{tex/appendix_2}
%\input{tex/appendix_3}
%\input{tex/worst-case-algebra}

\end{document}

%% file: cls/sym.tex
\newcommand{\calC}{{\cal C}}

\newcommand{\calE}{{\cal E}}
\newcommand{\calF}{{\cal F}}
\newcommand{\calG}{{\cal G}}

\newcommand{\calK}{{\cal K}}
\newcommand{\calL}{{\cal L}}

\newcommand{\calO}{{\cal O}}
\newcommand{\calP}{{\cal P}}

\newcommand{\calX}{{\cal X}}

\newcommand{\bfd}{\mathbf{d}}

\newcommand{\bfk}{\mathbf{k}}

\newcommand{\bfn}{\mathbf{n}}

\newcommand{\bfp}{\mathbf{p}}
\newcommand{\bfq}{\mathbf{q}}

\newcommand{\bfu}{\mathbf{u}}

\newcommand{\bfx}{\mathbf{x}}

\newcommand{\bftheta}{\boldsymbol{\theta}}

\newcommand{\bfI}{\mathbf{I}}

\newcommand{\bfQ}{\mathbf{Q}}
\newcommand{\bfR}{\mathbf{R}}

\newcommand{\bbN}{\mathbb{N}}

\newcommand{\bbR}{\mathbb{R}}

\newcommand{\ubfu}{\underline{\bfu}}

%% file: tex/Intro.tex
\section{Introduction}
\label{sec: intro}

% Operating safely in dynamic environments is crucial for autonomous robots in real-world scenarios. Existing control barrier functions for obstacle avoidance often assume point or circular robots, limiting their applicability to robots with more complex geometries. In this paper, we address this limitation by presenting an analytic approach to compute the distance between a polygonal robot and moving elliptical obstacles in a 2D environment. This distance computation is utilized in constructing a control barrier function for safe control synthesis, enabling the operation of a robot with a more intricate shape. Our proposed approach offers real-time tight elliptical obstacle avoidance for polygon-shaped robots. 

Obstacle avoidance in static and dynamic environments is a central challenge for safe mobile robot autonomy. 

At the planning level, several motion planning algorithms have been developed to provide a feasible path that ensures obstacle avoidance, including prominent approaches like A$^*$~\cite{A_star_planning}, RRT$^*$~\cite{RRT_star}, and their variants~\cite{informed_rrt_star, neural_rrt_star}. These algorithms typically assume that a low-level tracking controller can execute the planned path. However, in dynamic environments where obstacles and conditions change rapidly, reliance on such a controller can be limiting. A significant contribution to the field was made by Khatib \cite{potential-field}, who introduced artificial potential fields to enable collision avoidance during not only the motion planning stage but also the real-time control of a mobile robot. Later, Rimon and Koditschek \cite{navigation-function} developed navigation functions, a particular form of artificial potential functions that guarantees simultaneous collision avoidance and stabilization to a goal configuration.
%. These functions strive to ensure collision avoidance and stabilization towards a goal configuration simultaneously. 
% Meanwhile, Fox \cite{Fox1997TheDW} introduced the dynamics window concept, an influential approach to obstacle avoidance that proactively filters out unsafe control actions. 
In recent years, research has delved into the domain of trajectory generation and optimization, with innovative algorithms proposed for quadrotor safe navigation \cite{mellinger_snap_2011, zhou2019robust, tordesillas2019faster}. In parallel, the rise of learning-based approaches \cite{michels2005high, pfeiffer2018reinforced, loquercio2021learning} has added a new direction to the field, utilizing machine learning to facilitate both planning and real-time obstacle avoidance. Despite their promise, these methods often face challenges in dynamic environments and in providing safety guarantees.

In the field of safe control synthesis, integrating control Lyapunov functions (CLFs) and control barrier functions (CBFs) into a quadratic program (QP) has proven to be a reliable and efficient strategy for formulating safe stabilizing controls across a wide array of robotic tasks \cite{glotfelter2017nonsmooth, grandia_2021_legged, Long2022RAL}. While CBF-based methodologies have been deployed for obstacle avoidance \cite{srinivasan2020synthesis, Long_learningcbf_ral21, almubarak2022safety, abdi2023safe}, such strategies typically simplify the robot as a point or circle and assume static environments when constructing CBFs for control synthesis. Some recent advances have also explored the use of time-varying CBFs to facilitate safe control in dynamic environments \cite{he2021rule, molnar2022safety, hamdipoor2023safe}. However, this concept has yet to be thoroughly investigated in the context of obstacle avoidance for robots with complex shapes. For the safe autonomy of robot arms, Koptev \textit{et al}. \cite{Koptev2023_neural_joint_control} introduced a neural network approach to approximate the signed distance function of a robot arm and use it for safe reactive control in dynamic environments. In \cite{Hamatani2020arm}, a CBF construction formula is proposed for a robot arm with a static and circular obstacle. 
% A configuration-aware control approach for the robot arm was proposed in \cite{ding2022configurationaware} by integrating geometric restrictions with CBFs. 
Thirugnanam \textit{et al}. \cite{discrete_polytope_cbf} introduced a discrete CBF constraint between polytopes and further incorporated the constraint in a model predictive control to enable safe navigation. The authors also extended the formulation for continuous-time systems in \cite{polytopic_cbf} but the CBF computation between polytopes is numerical, requiring a duality-based formulation with non-smooth CBFs. 

\subsubsection*{Notations}

The sets of non-negative real and natural numbers are denoted $\bbR_{\geq 0}$ and $\bbN$. For $N \in \bbN$, $[N] := \{1,2, \dots N\}$. The orientation of a 2D body is denoted by $0 \leq \theta < 2\pi$ for counter-clockwise rotation. We denote the corresponding rotation matrix as 
% \begin{equation}
% \label{eq: rotation}
    $\bfR(\theta) = \begin{bmatrix} \cos \theta & -\sin \theta \\ \sin \theta & \cos \theta \end{bmatrix}.$
% \end{equation}
The configuration of a 2D rigid-body is described by position and orientation, and the space of the positions and orientations in 2D is called the special Euclidean group, denoted as $SE(2)$. Also, we use $\|\bfx\|$ to denote the $L_2$ norm for a vector $\bfx$ and $\otimes$ to denote the Kronecker product. The gradient of a differentiable function $V$ is denoted by $\nabla V$, and its Lie derivative along a vector field $f$ by $\calL_f V  = \nabla V \cdot f$. A continuous function $\alpha: [0,a)\rightarrow [0,\infty )$ is of class $\calK$ if it is strictly increasing and $\alpha(0) = 0$. A continuous function $\alpha: \bbR \rightarrow \mathbb{R}$ is of extended class $\calK_{\infty}$ if it is strictly increasing, $\alpha(0) = 0$, and $\lim_{r \rightarrow \infty} \alpha(r) = \infty$. Lastly, consider the body-fixed frame of the ellipse $\calE'$. The signed distance function (SDF) of the ellipse $\psi_\calE: \mathbb{R}^2 \to \mathbb{R}$ is defined as 
\begin{equation}
%\label{eq: SDF}
    \psi_\calE(\bfp') 
    = \left\{
    \begin{array}{ll}
        d(\calE',\bfp'), & \text{if } \bfp' \in \calE^c,  \\
        -d(\calE',\bfp'), & \text{if } \bfp' \in \calE,
    \end{array} 
    \right. \notag
\end{equation}
where $d$ is the Euclidean distance. In addition, $\|\nabla \psi_\calE (\bfp')\| = 1$ for all $\bfp'$ except on the boundary of the ellipse and its center of mass, the origin.

\textbf{Contributions}: (i) We present an analytic distance formula in $SE(2)$ for elliptical and polygonal objects, enabling closed-form calculations for distance and its gradient. (ii) We introduce a novel time-varying control barrier function, specifically for rigid-body robots described by one or multiple $SE(2)$ configurations. Its efficacy of ensuring safe autonomy is demonstrated in ground robot navigation and multi-link robot arm problems.

%% file: tex/ProblemFormulation.tex
\section{Problem Formulation}\label{sec:problem}

%\subsection{Notations}
% \footnotetext[1]{\textbf{Notation. }
% The sets of non-negative real and natural numbers are denoted $\bbR_{\geq 0}$ and $\bbN$. For $N \in \bbN$, $[N] := \{1,2, \dots N\}$. The orientation of a 2D body is denoted by $0 \leq \theta < 2\pi$ for counter-clockwise rotation. We denote the corresponding rotation matrix as 
% % \begin{equation}
% % \label{eq: rotation}
%     $\bfR(\theta) = \begin{bmatrix} \cos \theta & -\sin \theta \\ \sin \theta & \cos \theta \end{bmatrix}.$
% % \end{equation
% The $L_2$ norm for a vector $\bfx$ is denoted by $\|\bfx\|$. The gradient of a differentiable function $V$ is denoted by $\nabla V$, and its Lie derivative along a vector field $f$ by $\calL_f V  = \nabla V \cdot f$. A continuous function $\alpha: [0,a)\rightarrow [0,\infty )$ is of class $\calK$ if it is strictly increasing and $\alpha(0) = 0$. A continuous function $\alpha:\mathbb{R} \rightarrow \mathbb{R}$ is of extended class $\calK_{\infty}$ if it is of class $\calK$ and $\lim_{r \rightarrow \infty} \alpha(r) = \infty$.}

% \NA{The footnote should be referenced somewhere. It might be better to just introduce the notation in the main text whenever it is needed.}

Consider a robot with dynamics governed by a non-linear control-affine system,
\begin{equation}
\label{eq: dynamic}
\begin{aligned}
    &\dot{\bfx} = f(\bfx) + g(\bfx) \bfu ,
\end{aligned}
\end{equation}
where $\bfx \in \calX \subseteq \mathbb{R}^{n}$ is the robot state and $\bfu \in  \mathbb{R}^{m}$ is the control input. Assume that $f : \mathbb{R}^{n} \mapsto \mathbb{R}^{n}$ and $g : \mathbb{R}^{n} \mapsto \mathbb{R}^{n \times m}$ are continuously differentiable functions. We assume the robot operates in a 2D workspace with a state-dependent shape $S(\bfx) \subset \bbR^2$.

% \NA{Move this paragraph later, now that we define the robot shape $S(\bfx)$, it is not necessary to define the polygon here. It may be better to use a different letter than $S$ since (a) we use calligraphic fonts for sets and (b) we are defining $\calS$ as the safe region. One idea is to define $\calS(\bfx)$ as the robot shape and $\calF$ as the free space.}

We assume the $\bbR^2$ workspace is partitioned into a closed safe (free) region $\mathcal{F}(t)$ and an open unsafe region $\mathcal{O}(t)$ such that  $\mathcal{F}(t) \cap \mathcal{O}(t) = \emptyset$ and $\bbR^2 = \mathcal{F}(t) \cup \mathcal{O}(t)$. We assume the unsafe set $\mathcal{O}(t)$ is characterized by a collection of dynamical elliptical obstacles with known rigid-body motions, denoted as $\{\calE(\bfq_i(t), \bfR(\theta_i(t)), a_i, b_i)\}_{i=1}^N$. Here, $\bfq_i$ denotes the center of mass and $\bfR_i$ denotes the rotation matrix of the ellipse. In its body-fixed frame, $a_i$ and $b_i$ are the lengths of the semi-axes of the ellipse along the $x$-axis and $y$-axis, respectively.

\begin{problem*}
Given a robot with shape $S(\bfx)$ governed by dynamics \eqref{eq: dynamic} that can perfectly determine its state, the objective is to stabilize the robot safely within a goal region $\calG \subset \calF(t) \; \forall t \geq 0$ such that $S(\bfx(t)) \cap \calO(t) = \emptyset$ for all $t \geq 0$.
\end{problem*}

%% file: tex/Prelim.tex
\section{Preliminaries}
\label{sec: prelim}

In this section, we review preliminaries on control Lyapunov and barrier functions and discuss their use in synthesizing a safe stabilizing controller for dynamics in~\eqref{eq: dynamic}.

\subsection{Control Lyapunov Function}
The notion of a control Lyapunov function (CLF) was introduced in \cite{Artstein1983StabilizationWR, SONTAG1989117} to verify the stabilizability of control-affine systems \eqref{eq: dynamic}. Specifically, a (exponentially stabilizing) CLF $V: \calX \mapsto \bbR$ is defined as follows,
\begin{definition}
A function $V \in \mathbb{C}^1(\calX,\mathbb{R})$ is a \emph{control Lyapunov function (CLF)} on $\calX$ for system \eqref{eq: dynamic} if $V(\bfx)>0, \forall \bfx \in \calX \setminus \{\boldsymbol{0} \}, V(\boldsymbol{0}) = 0$, and it satisfies:
\begin{equation}\label{eq: clf}
    \inf_{\bfu \in \bbR^m} \text{CLC}(\bfx,\bfu) \leq 0, \quad \forall \bfx \in \calX,
\end{equation}
where $\text{CLC}(\bfx,\bfu) := \mathcal{L}_f V(\bfx) + \mathcal{L}_g V(\bfx)\bfu + \alpha_V( V(\bfx))$
is the \emph{control Lyapunov condition} (CLC) defined for some class $K$ function $\alpha_V$.
\end{definition}

\subsection{Control Barrier Function}

To facilitate safe control synthesis, we consider a time-varying set $\calC(t)$ defined as the super zero-level set of a continuously differentiable function $h: \calX \times \bbR_{\geq 0} \mapsto \bbR$:
\begin{equation}
\label{eq: safe_set}  
    \calC(t) := \{\bfx \in \calX \subseteq \bbR^n: h(\bfx, t) \geq 0 \}.
\end{equation}
Safety of the system \eqref{eq: dynamic} can then be ensured by keeping the state $\bfx$ within the safe set $\calC(t)$.

\begin{definition}
\label{def: tv_cbf}
A function $h: \mathbb{R}^n \times \bbR_{\geq 0} \mapsto {\mathbb{R}}$ is a valid time-varying \emph{control barrier function (CBF)} on $\mathcal{X} \subseteq \mathbb{R}^n$ for \eqref{eq: dynamic} if there exists an extended class $\mathcal{K}_{\infty}$ function $\alpha_h$ with:
\begin{equation}\label{eq:tv_cbf}
    \sup_{\bfu\in \mathcal{U}} \text{CBC}(\bfx,\bfu, t) \geq 0, \quad \forall \; (\bfx,t) \in \calX \times \bbR_{\geq 0},
\end{equation}
where the \emph{control barrier condition (CBC)} is:
\begin{equation}
\label{eq:tvcbc_define}
\begin{aligned}
    &\text{CBC}(\bfx,\bfu, t) := \dot{h}(\bfx, t) + \alpha_h(h(\bfx,t)) \\
    & = \mathcal{L}_f h(\bfx, t) + \mathcal{L}_g h(\bfx, t)\bfu + \frac{\partial h(\bfx,t)}{\partial t} + \alpha_h(h(\bfx,t)).
\end{aligned}
\end{equation}
\end{definition}

Suppose we are given a baseline feedback controller $\bfu = \bfk(\bfx)$ for the control-affine systems \eqref{eq: dynamic}, and we aim to ensure the safety and stability of the system. By observing that both the CLC and CBC constraints are affine in the control input $\bfu$, a quadratic program (QP) can be formulated for online synthesis of a safe stabilizing controller for \eqref{eq: dynamic}:
\begin{equation}
\begin{aligned}
\label{eq: clf_cbf_qp}
    \bfu(\bfx) = &\argmin_{\bfu \in \bbR^m,\delta \in \bbR} \| \bfu - \bfk(\bfx)\|^2 + \lambda \delta^2,  \\
    \mathrm{s.t.} \, \,  &\text{CLC}(\bfx,\bfu) \leq \delta,  \text{CBC}(\bfx,\bfu, t) \geq 0,
\end{aligned}
\end{equation}
where $\delta \geq 0$ denotes a slack variable that relaxes the CLF constraints to ensure the feasibility of the QP, controlled by the scaling factor $\lambda > 0$.

%% file: tex/Polygon_Ellipse_Env.tex
\section{Analytic distance between ellipse and polygon}
\label{sec: analytic_distance}

In this section, we derive an analytic formula for computing the distance between a polygon and an ellipse, which enables the formulation of CBFs to ensure safe autonomy.

% For notation, the orientation of a 2D body is given by the parameter $0 \leq \theta < 2\pi$ for counter-clockwise rotation. For convenience, it is also expressed as 
% \begin{equation}
% \label{eq: rotation}
%     \bfR(\theta) = \begin{bmatrix} \cos \theta & -\sin \theta \\ \sin \theta & \cos \theta \end{bmatrix},
% \end{equation}
% and $\bfR$ might be written instead when $\theta$ is understood.

% For an arbitrary elliptical obstacle $\calE(\bfq, \bfR(\theta), a, b)$ in the inertial frame, $\bfq$ is the center of mass, and $\bfR$ denotes the rotation matrix of the ellipse. In its body-fixed frame, $a$ and $b$ are the lengths of the semi-axes of the ellipse along the $x$-axis and $y$-axis, respectively. Similarly, let $\calP( \tilde{\bfq}, \tilde{\bfR}(\tilde{\theta}), \{\tilde{\bfp}_i\}_{i=0}^{M-1})$ denote the polygon of interest where $\Tilde{\bfq}$ as the center of mass and $\Tilde{\bfR}$ as the orientation in the inertial frame. In its fixed-body frame, $\{\tilde{\bfp}_i\}$ are the vertices of the polygonal robot with line segments $\tilde{\bfd}_i = \tilde{\bfp}_{[i+1]_M} - \tilde{\bfp}_i$ for $i = 0, 1, \ldots, M-1$ where $[\cdot]_M$ is the $M$-modulus. 

We consider the mobile robot's body $S(\bfx)$ to be described as a polygon, denoted by $\calP( \tilde{\bfq}, \tilde{\bfR}(\tilde{\theta}), \{\tilde{\bfp}_i\}_{i=0}^{M-1})$. Here, $\Tilde{\bfq}$ denotes the center of mass and $\Tilde{\bfR}$ denotes the orientation in the inertial frame. In its fixed-body frame, $\{\tilde{\bfp}_i\}$ denotes the vertices of the robot with line segments $\tilde{\bfd}_i = \tilde{\bfp}_{[i+1]_M} - \tilde{\bfp}_i$ for $i = 0, 1, \ldots, M-1$ where $[\cdot]_M$ is the $M$-modulus. 

For convenience, denote $\calE$ and $\calP$ as the bodies in the inertial frame, and we assume their intersection is empty. Now, denote $\calE'$ and $\calP'$ as the respective bodies in the body-fixed frame of the elliptical obstacle. As a result, 
%\begin{equation}
%\label{eq: distance_frames}
    $d(\calE, \calP) = d(\calE',\calP')$
%\end{equation}
by isometric transformation.

Furthermore, let $\tilde{\bfp}_i$ be a vertex in the robot's frame. Then in the inertial frame, it becomes $\bfp_i = \Tilde{\bfq} + \tilde{\bfR} \Tilde{\bfp}_i$. In the obstacle's frame, it is 
\begin{align}
    \bfp_i' 
        = \bfR^\top(\bfp_i - \bfq)
        = \bfR^\top \tilde{\bfR} \Tilde{\bfp}_i + \bfR^\top(\Tilde{\bfq} -\bfq),
\label{eq: vertices_ellipse_frame}
\end{align}
In short, $\{\tilde{\bfp}_i\}$ are vertices in the robot's frame, $\{\bfp_i\}$ are vertices in the inertial frame, and $\{\bfp_i'\}$ are vertices in the obstacle's frame. The distance function is
\begin{equation}
\label{eq: polygon_ellipse_1}
        d(\calE', \calP') := \min_{i \in [M-1]} d(\calE', \bfd_i'),
\end{equation}
which computes the distance between the ellipse $\calE'$ and each line segment $\bfd_i'$. We write each segment as
\begin{equation}
\label{eq: line_segment_i}
    l_i'(\tau) = (1-\tau)\bfp_i' + \tau \bfp_{[i+1]_{M}}',
\end{equation}
for $\tau \in [0,1]$. This further simplifies the function to
\begin{equation}
\label{eq: polygon_line_seg_1}
    d(\calE', \bfd_i') = \min_{\tau \in [0,1]} d(\calE', l_i'(\tau)).
\end{equation}

Now, there are essentially two groups of computations for the distance in \eqref{eq: polygon_line_seg_1}: one is the distance between the ellipse $\calE'$ and the endpoints of $\bfd_i'$; the other is the distance between the ellipse $\calE'$ and the infinite line $l_i'(\tau)$ for arbitrary $\tau$ with the caveat that the minimizing argument occurs at $\tau^* \in (0,1)$. The two computations are detailed in the procedures which follow our next proposition.

\begin{proposition}
    Let $\calE'$ be an ellipse and $l_i'$ be a line segment in the frame of the ellipse. Denote $\tau^*$ as the argument of the minimum in~\eqref{eq: polygon_line_seg_1}. Then, the distance
    \begin{equation}
        d(\calE', \bfd_i') = 
        \left\{
            \begin{array}{ll}
            \| \bfp_i' - \underline{\bfp_i'} \|,     &  \text{if } \tau^* = 0, \\
            \| \bfp_{[i+1]_{M}}' - \underline{\bfp_{[i+1]_{M}}'} \|,      & \text{if } \tau^* = 1,  \\
            \| l_i'(\tau^*) - \underline{l_i'(\tau^*)} \|,    & \text{if } \tau^* \in (0,1),
            \end{array}
        \right.
    \end{equation}
where $\underline{\mathbf{p}i'}$ and $\underline{\mathbf{p}{[i+1]{M}}'}$ are the points on the ellipse closest to $\mathbf{p}i'$ and $\mathbf{p}{[i+1]{M}}'$, respectively. These points are determined using \textbf{Procedure 1}. The terms $l_i'(\tau^*)$ and $\underline{l_i'(\tau^*)}$ (on the ellipse) are determined using \textbf{Procedure 2}.
\label{prop: distance}
\end{proposition}

\begin{method}
Let $\bfp' = (p_x', p_y')$ be one of the endpoints for the line segment $\bfd_i'$. Recall that the ellipse is defined by its semi-axes along $x$-axis and $y$-axis, denoted by $a$ and $b$, respectively. The points on the ellipse are parameterized by 
\begin{equation}
x(t) = a\cos(t), \quad y(t) = b\sin(t),
\end{equation}
for $0 \leq t \leq 2 \pi$. The goal is to determine the point $(x(t), y(t))$ on the ellipse that is closest to the point $\bfp'$, so it is a minimization problem of the squared Euclidean distance:
\begin{equation}
d^2(t) = (p_x' - a\cos(t))^2 + (p_y' - b\sin(t))^2.
\end{equation}
To find the minimum distance, we determine the critical point(s) by solving for $0 = \frac{d}{dt} d^2(t)$, which simplified to
\begin{equation}
0 = (b^2 - a^2)\cos t \sin t  + a p_x' \sin t - b p_y' \cos t.
\end{equation}
Using single-variable optimization, we substitute 
\begin{equation}
    \cos t = \lambda, \quad \sin t = \sqrt{1-\lambda},
\end{equation}
and this yields $b p_y' \lambda = \sqrt{1-\lambda^2}((b^2-a^2)\lambda+a p_x')$, which is a quartic equation in $\lambda$. Furthermore, a monic quartic can be derived, which gives the following simplified coefficients:
\begin{equation}
    0 = \lambda^4 + 2m\lambda^3 + (m^2 + n^2 -1) \lambda^2 -2m \lambda -m^2,
\end{equation}
where
\begin{equation}
    m = p_x' \frac{a}{b^2 - a^2}, \quad n = p_y' \frac{b}{b^2 - a^2}.
\end{equation}
From this point, the real root(s) of the equation can be solved analytically following Cardano's and Ferrari's solution for the quartic equations \cite{cardano2011artis}. 
% (The solution is coded\NA{This is informal. The right word is implemented. Is it necessary to provide this reference?} in \cite{ellipsesdf} with further simplifications using the symmetries of the coefficients.) 
Let $\underline{t}$ be the solution so that $\underline{\bfp'} = (x(\underline{t}),y(\underline{t}))$ is a point on the ellipse and is closest to $\bfp'$. Hence, 
\begin{equation}
\label{eq: ellipse_to_point}
    d(\calE', \bfd_i') = \| \bfp' - \underline{\bfp'} \|
\end{equation}
where $\bfp'$ is either $\bfp_i'$ or $\bfp_{[i+1]_{M}}'$. 
% This concludes \textbf{Procedure 1}.
\end{method}

\begin{method}
We compute the distance between the ellipse $\calE'$ and the infinite line $l_i'(\tau)$ whose minimizing point occurs at $\tau^* \in (0,1)$. First, define the unit normal of the infinite line as
\begin{equation}
\label{eq: unit_normal_di}
    \hat{\bfn}_i' = \frac{1}{\| \bfd_i' \|}(-d'_{i,y},d'_{i,x}).
\end{equation}
Denote $\underline{l_i'(\tau^*)}$ as the point on the ellipse that is closest to the $l_i'(\tau^*)$. In fact, this point $\underline{l_i'(\tau^*)}$ must have a tangent line at the ellipse which is parallel to $l_i'$; which means the normal at $\underline{l_i'(\tau^*)}$ is $\pm \hat{\bfn}_i'$. Therefore, we compute the point on the ellipse up to a sign:
\begin{equation}
\label{eq:closest_point_to_line}
    \underline{l_i'(\tau^*)} = \pm \frac{I_\epsilon^2 \hat{\bfn}_i'}{\| I_\epsilon \hat{\bfn}_i' \|},
\end{equation}
where $I_\epsilon = \text{diag}(a,b)$. The correct sign is chosen when we are looking at the sign of the constant $C$ in the line equation $Ax + By + C = 0$ of $l_i'$. In particular, 
\begin{equation}
\label{eq:line_equation_constant}
    C = -\hat{\bfn}_i'^\top \bfp_i'.
\end{equation}
If $C > 0$, then $\underline{l_i'(\tau^*)} = -\frac{I_\epsilon^2 \hat{\bfn}_i'}{\| I_\epsilon \hat{\bfn}_i' \|}$; otherwise, if $C < 0$, then $\underline{l_i'(\tau^*)} =  \frac{I_\epsilon^2 \hat{\bfn}_i'}{\| I_\epsilon \hat{\bfn}_i' \|}$. Finally, we determine $l_i'(\tau^*)$ on the line segment $\bfd_i'$ using projection:
\begin{equation}
\label{eq: ellipse_to_line_closestPoint}
    l_i'(\tau^*) = \bfp_i' + \text{proj}_{\bfd_i'} (\underline{l_i'(\tau^*)} - \bfp_i').
\end{equation}
Here, we are done with \textbf{Procedure 2}. 
\end{method}

Next, we compute the partial derivatives of 
$d(\calE', \calP')$ with respect to either $(\bfq,\bfR)$, the configuration of the obstacle, or $(\tilde{\bfq}, \tilde{\bfR})$, the configuration of the polygonal robot.

In general, both procedures above compute the distance using the Euclidean norm between two unique points: one point $\bfp'$ on a line segment of the robot, and the other $\underline{\bfp'}$ on the ellipse. This is, in fact, equivalent to the SDF of the ellipse evaluated at $\bfp'$ by the uniqueness of these two points. Therefore, let $\bfp' = l_i(\tau^*)$ for some $0 \leq i < M$, then 
\begin{equation}
    d(\calE', \calP') = \psi_\calE(\bfp') = \psi_\calE(l_i(\tau^*)).
\end{equation}
Then, its gradient with respect to $\bfp'$ is
%\begin{equation}
%\label{eq: ellipse_to_point_gradient}
    $\nabla \psi_\calE(\bfp') = \frac{\bfp' - \underline{\bfp'}}{\| \bfp' - \underline{\bfp'} \|} $. 
%\end{equation}
However, note that $\bfp'$ is a point transformed from the polygonal robot's frame using \eqref{eq: vertices_ellipse_frame}, which depends on the configurations of the elliptical obstacle and the robot. Hence the partial derivatives can be computed as follows.
\begin{proposition}
    Let $\calE'$ and $\calP'$ be the elliptical obstacle and polygonal robot, respectively, in the obstacle's frame. Let $\bfp'$ and $\underline{\bfp'}$ be determined from Proposition \ref{prop: distance}, then 
    \begin{align}
    \frac{\partial d}{\partial \bfq} 
    &= \left( \frac{\partial d}{\partial q_x},\frac{\partial d}{\partial q_y} \right)
    = -\bfR \nabla \psi_\calE(\bfp') \label{eq: ellipse_to_point_partial_1}, \\
    \frac{\partial d}{\partial \bfR} 
    &= \nabla \psi_\calE(\bfp')\otimes (\tilde{\bfR} \tilde{\bfp} + (\tilde{\bfq} - \bfq))\label{eq: ellipse_to_point_partial_2}, \\
    \frac{\partial d}{\partial \tilde{\bfq}} 
    &= \left( \frac{\partial d}{\partial \tilde{q}_x},\frac{\partial d}{\partial \tilde{q}_y} \right)
    = \bfR \nabla \psi_\calE(\bfp') \label{eq: ellipse_to_point_partial_3}, \\
    \frac{\partial d}{\partial \tilde{\bfR}} 
    &= \bfR (\nabla \psi_\calE(\bfp') \otimes \tilde{\bfp}) \label{eq: ellipse_to_point_partial_4}.
\end{align}
Furthermore, \eqref{eq: ellipse_to_point_partial_2} and \eqref{eq: ellipse_to_point_partial_4} are derivatives with respect to the rotation matrices; one may compute the derivatives with respect to the rotation angle as
\begin{align}
    \begin{split}
    \frac{\partial d}{\partial \theta} 
        &= \nabla \psi_\calE(\bfp')^\top \left[\frac{\partial \bfR}{\partial \theta}^\top (\tilde{\bfR} \tilde{\bfp} + (\tilde{\bfq} - \bfq)) \right] \\
        &= \text{tr}\left[ \frac{\partial d}{\partial \bfR} \frac{\partial \bfR}{\partial \theta} \right],
    \end{split} \label{eq: ellipse_to_point_partial_2_v2} \\
    \begin{split}
    \frac{\partial d}{\partial \tilde{\theta}}  
        &= \nabla \psi_\calE(\bfp')^\top \left[ \bfR^\top \frac{\partial \tilde{\bfR}}{\partial \tilde{\theta}} \tilde{\bfp} \right] 
        = \text{tr}\left[ \frac{\partial d}{\partial \tilde{\bfR}} \frac{\partial \tilde{\bfR}}{\partial \tilde{\theta}}^\top \right].
    \end{split}
\end{align}
\end{proposition}

\begin{comment}
Consequently, the partial derivatives will be the same as equations \eqref{eq: ellipse_to_point_partial_1}--\eqref{eq: ellipse_to_point_partial_4}, where 
\begin{equation}
\label{eq: ellipse_to_point_gradienSpecial}
    \nabla \psi_\calE(\bfp_{\Delta_i}') = \frac{\bfp_{\Delta_i}' - \underline{\bfp_{\Delta_i}'}}{\| \bfp_{\Delta_i}' - \underline{\bfp_{\Delta_i}'} \|},
\end{equation}
and $\tilde{\bfp}_{\Delta_i} = \tilde{\bfR}^\top \bfR \bfp_{\Delta_i}'  + \tilde{\bfR}^\top(\bfq -\tilde{\bfq})$ in the robot's frame.
\end{comment}

Following both propositions above, we compute the distance function 
\begin{equation}
\label{eq: distance_ellipse_polygon_final}
    \Phi(\bfq, \bfR, \tilde{\bfq}, \tilde{\bfR}) = d(\calE, \calP) = d(\calE',\calP')
\end{equation}
for the elliptical obstacle $\calE(\bfq, \bfR, a, b)$ and the polygonal robot $\calP(\tilde{\bfq}, \tilde{\bfR}, \{\tilde{\bfp}_i\})$.

%% file: tex/MultipleObstacle.tex
\section{Polygon-Shaped Robot Safe Navigation in Dynamic Ellipse Environments}
\label{sec: multiple_moving_ellipses_env}

In this section, using the distance formula in \eqref{eq: distance_ellipse_polygon_final} and assuming the known motion of the ellipse obstacles, we derive TV-CBFs to ensure safety for polygon-shaped robots operating in dynamic elliptical environments. 

\subsection{Time-Varying Control Barrier Function Constraints}

We assume that there are a total of $N$ elliptical obstacles in the environment, each having a rigid-body motion with linear velocity $v_i$ and angular velocity $\omega_i$ around its center of mass. We define each time-varying CBF as 
\begin{equation}
\label{eq: tv_cbf_define}
    h_i(\bfx, t) := \Phi(\bfq_i(t), \bfR_i(t), \tilde{\bfq}, \tilde{\bfR}),
\end{equation}
where $\Phi$ is the collision function  and $\bfq_i(t)$ and $\bfR_i(t)$ denotes the position and orientation of the $i$-th ellipse at time $t$.

\begin{figure*}[h]
    \centering
    % First row
    \includegraphics[width=0.19\textwidth, trim={0.5cm 0.2cm 2cm 0.0cm},clip]{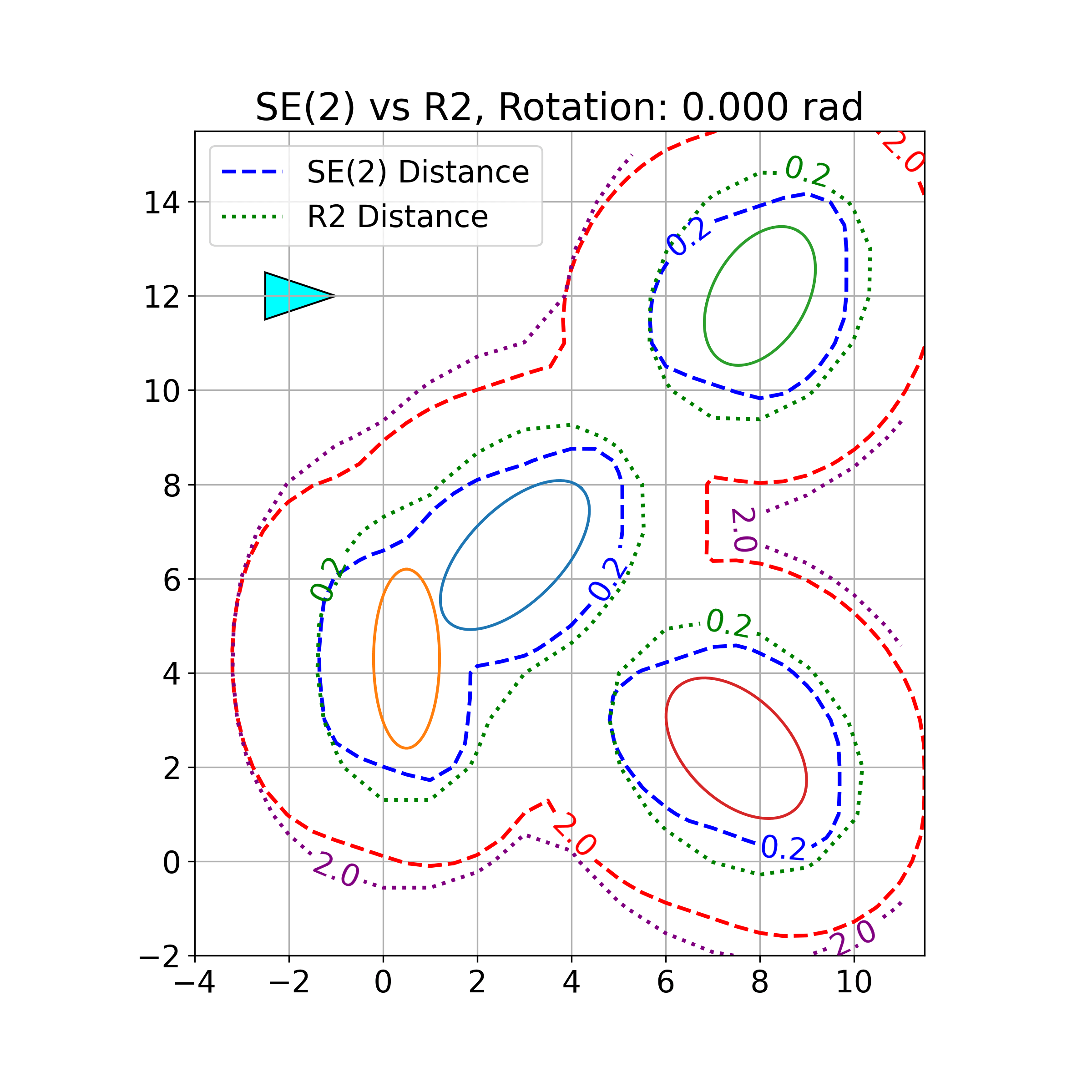}
    \includegraphics[width=0.19\textwidth, trim={0.5cm 0.2cm 2cm 0.0cm},clip]{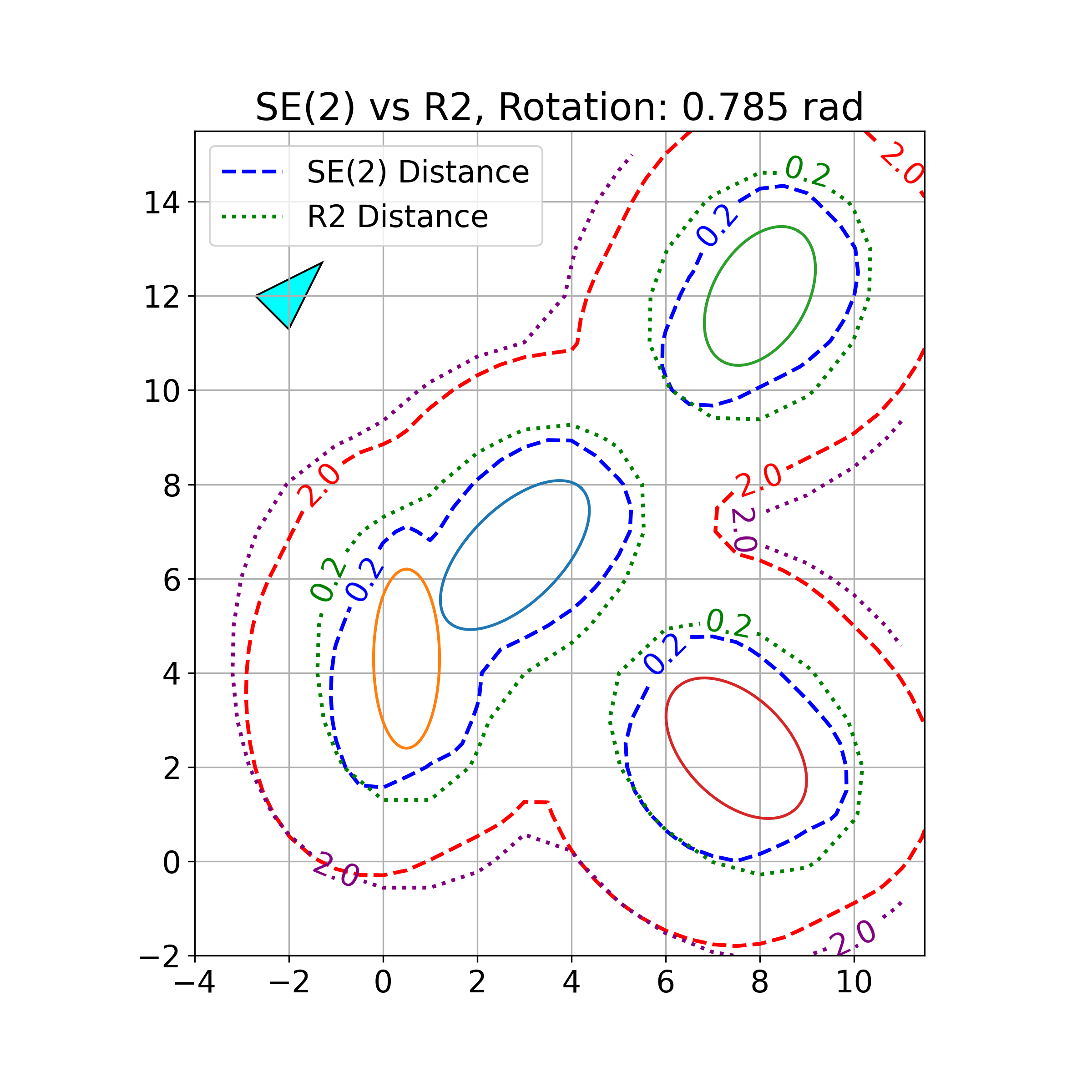}
    \includegraphics[width=0.19\textwidth, trim={0.5cm 0.2cm 2cm 0.0cm},clip]{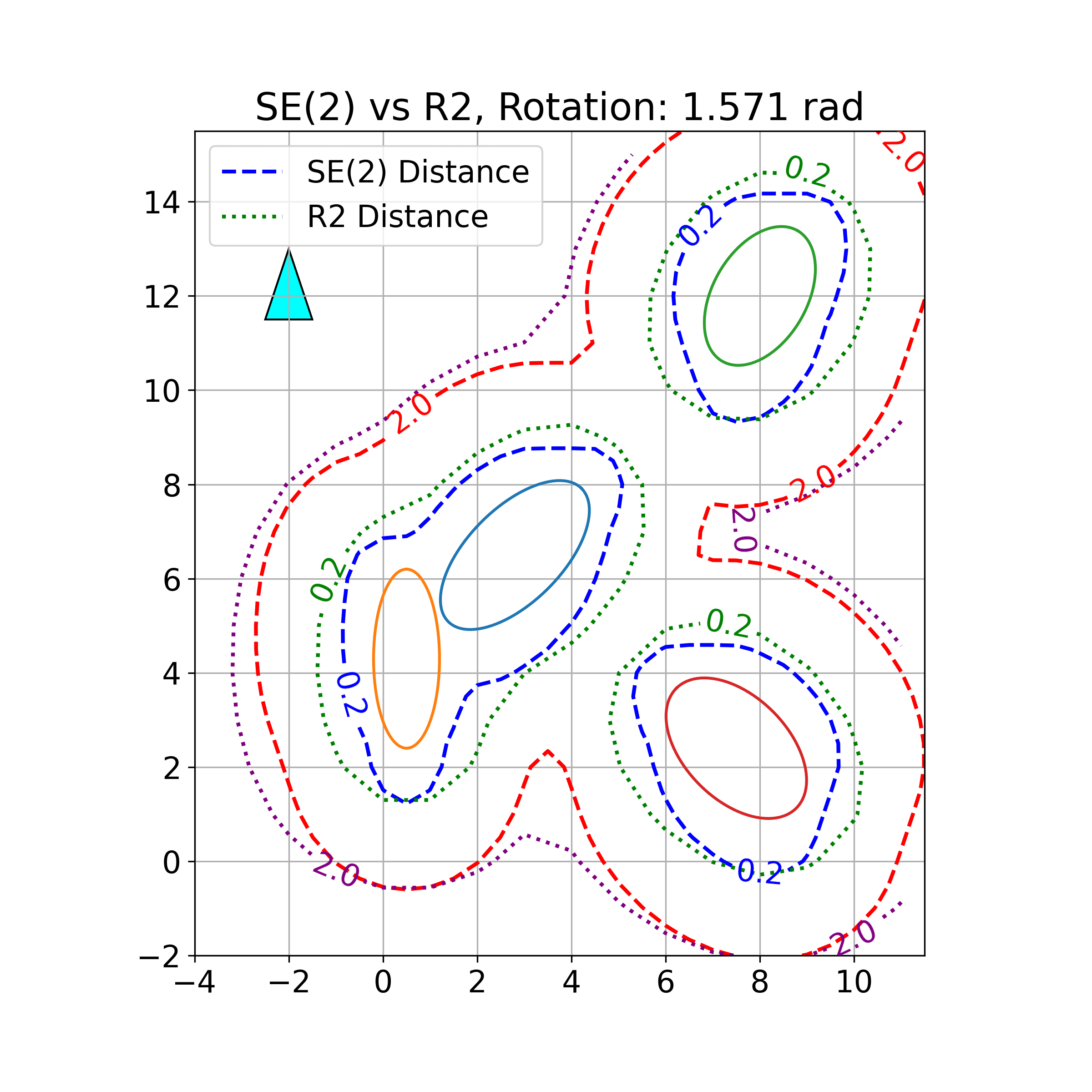}
    \includegraphics[width=0.19\textwidth, trim={0.5cm 0.2cm 2cm 0.0cm},clip]{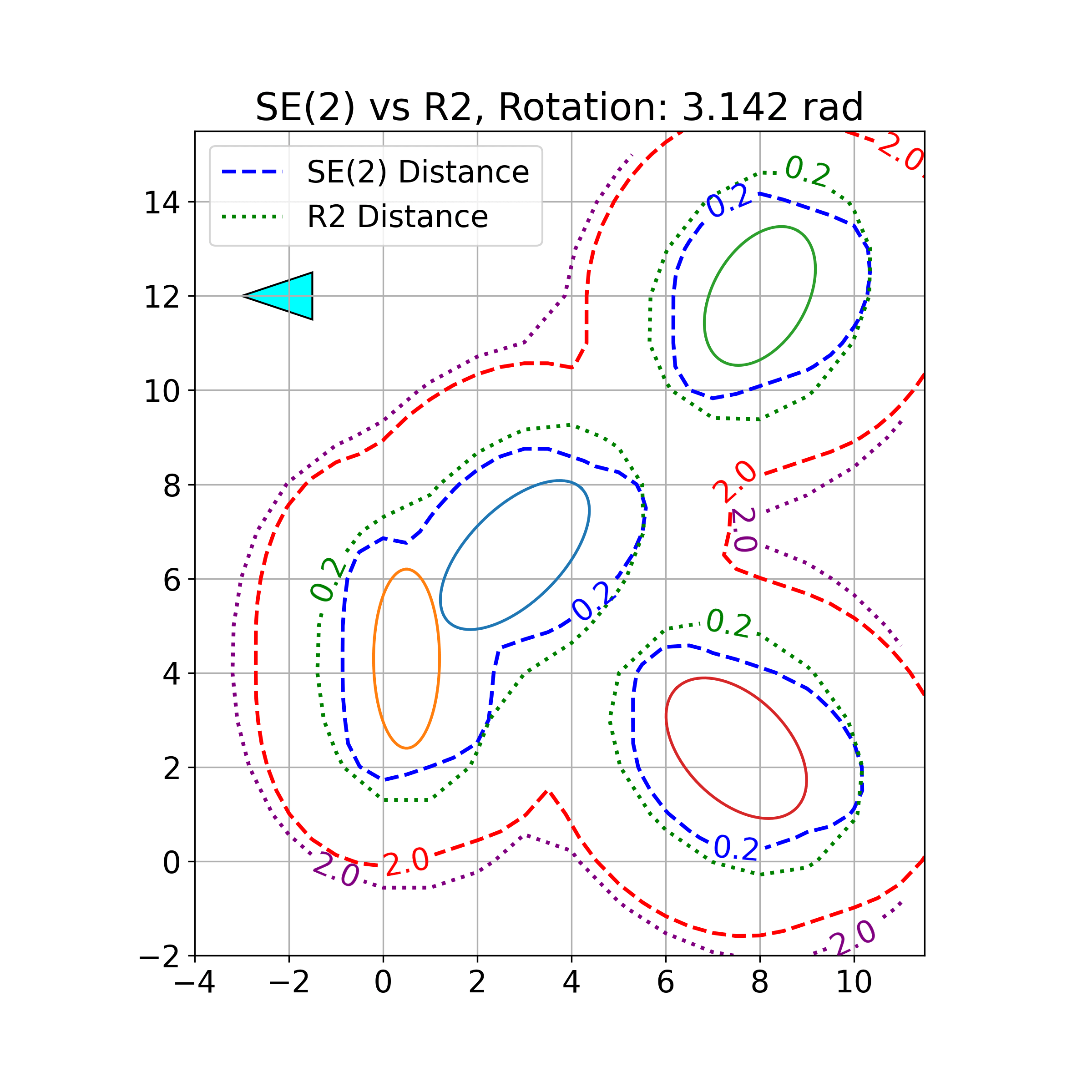}
    \includegraphics[width=0.19\textwidth, trim={0.5cm 0.2cm 2cm 0.0cm},clip]{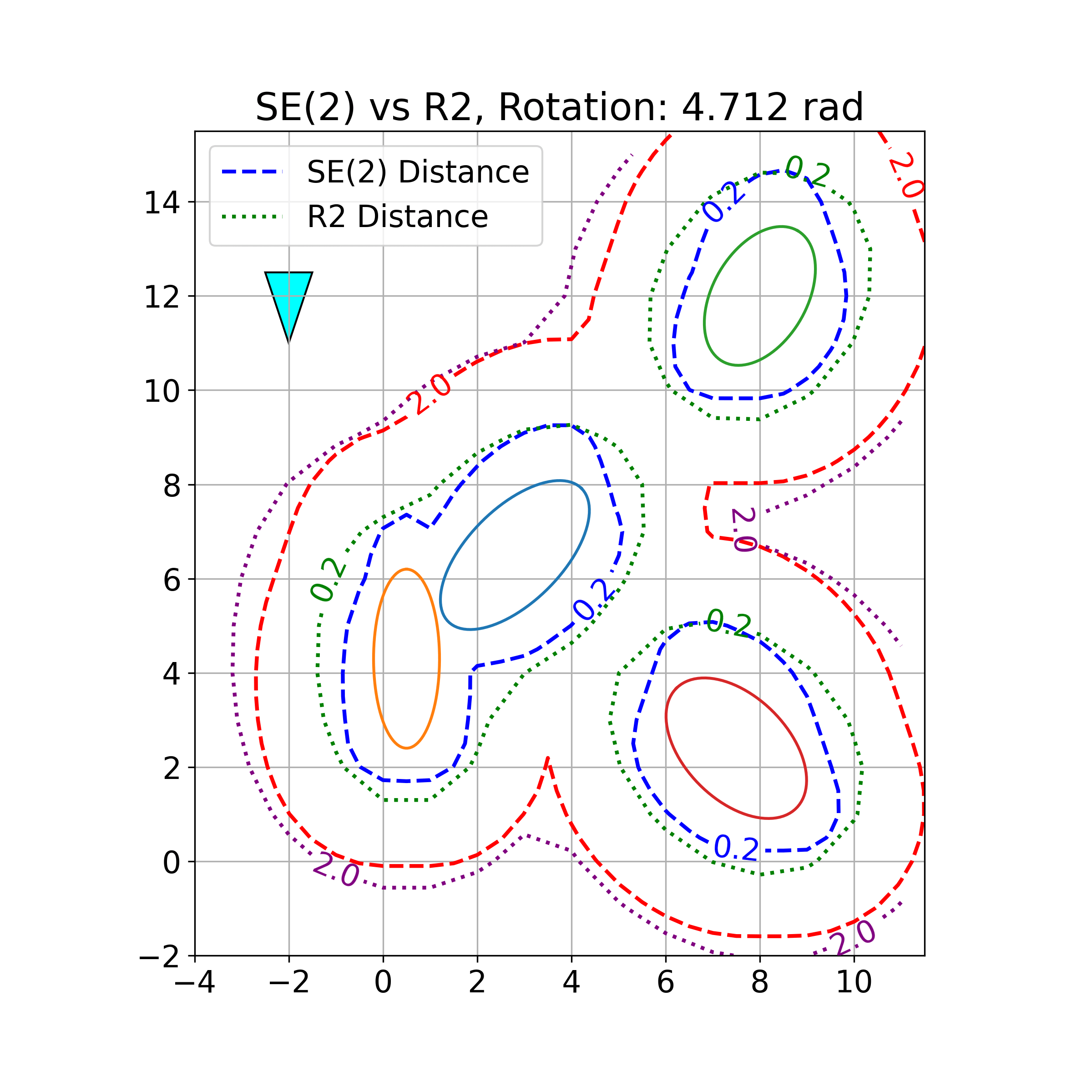}
    \caption{Comparative analysis of the $SE(2)$ and $\bbR^2$ signed distance functions for elliptical obstacles. The cyan triangle represents the rigid-body robot, with its orientation varying across the sequence. The importance of considering robot orientation in distance computations becomes evident: while the $SE(2)$ function accounts for this orientation, the $\bbR^2$ approximation treats the robot as an encapsulating circle with radius $1$. Level sets at distances $0.2$ and $2$ are depicted for both functions.}
    \label{fig:se2_r2_compare}
\end{figure*}

% \begin{figure*}[h]
%     \centering
%     % First row
%     \includepdf[pages={1}, scale=0.5, pagecommand={\includegraphics[width=0.19\textwidth, trim={0.5cm 0.2cm 2cm 0.0cm},clip]{fig/SE(2)_R2_compare/distance_plot_00.PNG}}]{}
%     \includepdf[pages={1}, scale=0.5, pagecommand={\includegraphics[width=0.19\textwidth, trim={0.5cm 0.2cm 2cm 0.0cm},clip]{fig/SE(2)_R2_compare/distance_plot_79.PNG}}]{}
%     \includepdf[pages={1}, scale=0.5, pagecommand={\includegraphics[width=0.19\textwidth, trim={0.5cm 0.2cm 2cm 0.0cm},clip]{fig/SE(2)_R2_compare/distance_plot_157.PNG}}]{}
%     \includepdf[pages={1}, scale=0.5, pagecommand={\includegraphics[width=0.19\textwidth, trim={0.5cm 0.2cm 2cm 0.0cm},clip]{fig/SE(2)_R2_compare/distance_plot_314.PNG}}]{}
%     \includepdf[pages={1}, scale=0.5, pagecommand={\includegraphics[width=0.19\textwidth, trim={0.5cm 0.2cm 2cm 0.0cm},clip]{fig/SE(2)_R2_compare/distance_plot_471.PNG}}]{}
%     \caption{Comparative analysis of the $SE(2)$ and $\bbR^2$ signed distance functions for elliptical obstacles. The cyan triangle represents the rigid-body robot, with its orientation varying across the sequence. The importance of considering robot orientation in distance computations becomes evident: while the $SE(2)$ function accounts for this orientation, the $\bbR^2$ approximation treats the robot as an encapsulating circle with radius $1$. Level sets at distances $0.2$ and $2$ are depicted for both functions.}
%     \label{fig:se2_r2_compare}
% \end{figure*}

Now, by utilizing the known motion of the ellipses with linear and angular velocity $v_i$ and $\omega_i$, we can express the CBC condition as:
\begin{equation*}
\label{eq:tvcbc_explicit}
\begin{aligned}
\textit{CBC}_i(\bfx,\bfu, t) & :=
\left[\frac{\partial \Phi(\bfq_i, \bfR_i, \bfx)}{\partial \bfx} \right]^\top F(\bfx)\ubfu + \frac{\partial \Phi(\bfq_i, \bfR_i)}{\partial \bfq_i} v_i \\
&+ \frac{\partial \Phi(\bfq_i, \bfR_i)}{\partial \bfR_i} \frac{\partial \bfR_i}{\partial \theta_i} \omega_i + \alpha_h(\Phi(\bfq_i, \bfR_i, \bfx)) \geq 0.
\end{aligned}
\end{equation*}
%
% Finally, the safety constraint is formulated by ensuring that the CBC condition holds for all $(\bfx, t) \in \calX \times [t_0, t_1]$:
% \begin{equation}\label{eq:tv_cbf}
% \sup_{\bfu\in \mathcal{U}} \textit{CBC}(\bfx,\bfu, t) \geq 0, \quad \forall \; (\bfx,t) \in \calX \times [t_0, t_1].
% \end{equation}

\subsection{Ground-Robot Navigation}

\begin{figure*}[h]
    \centering
    % First row
    \subcaptionbox{Initial Pose \label{fig:2a}}
    {\includegraphics[width=0.19\textwidth, trim={3cm 0.2cm 3cm 0.0cm},clip]{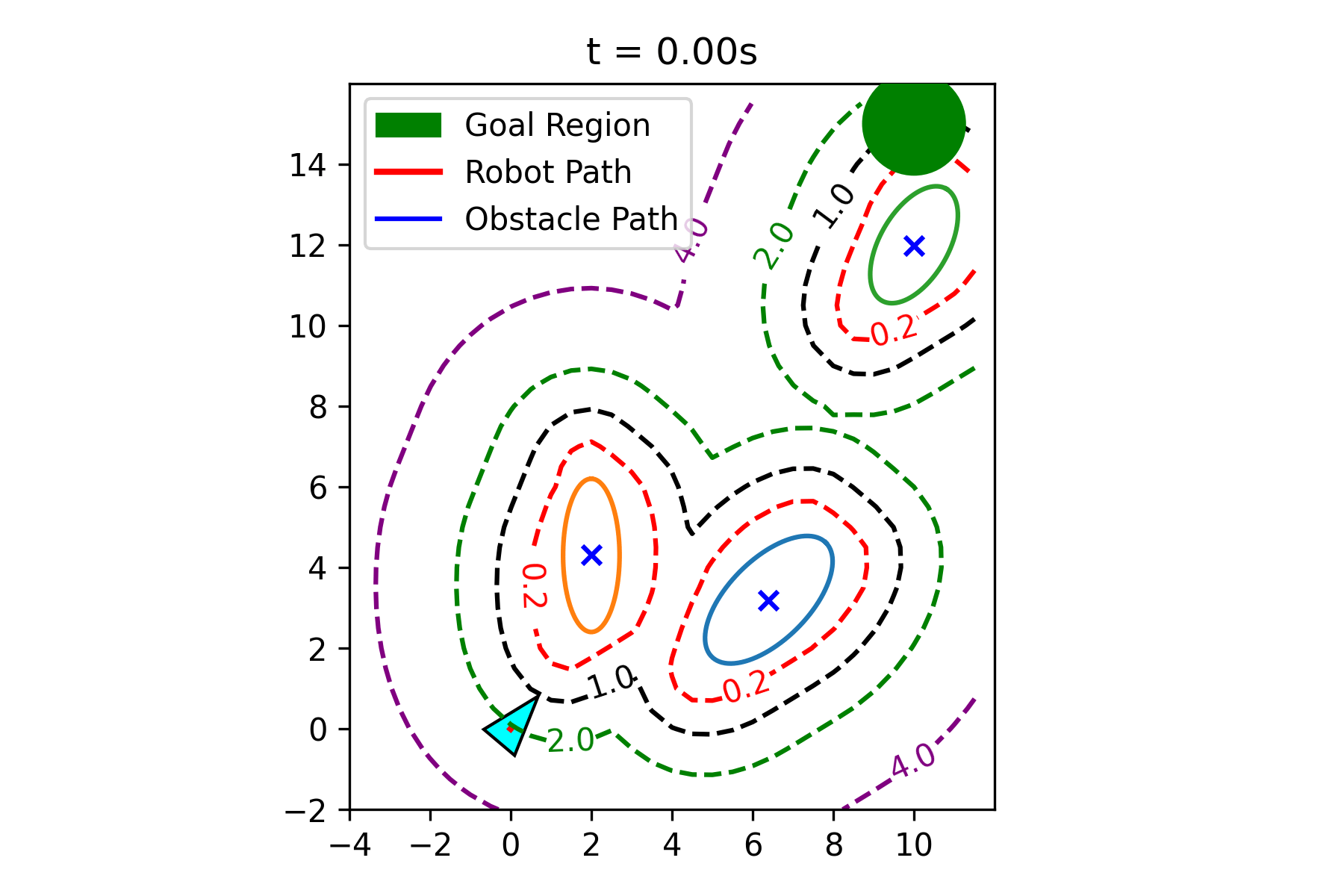}}
    \subcaptionbox{Time t = 1.66 sec \label{fig:2b}}
    {\includegraphics[width=0.19\textwidth, trim={3cm 0.2cm 3cm 0.0cm},clip]{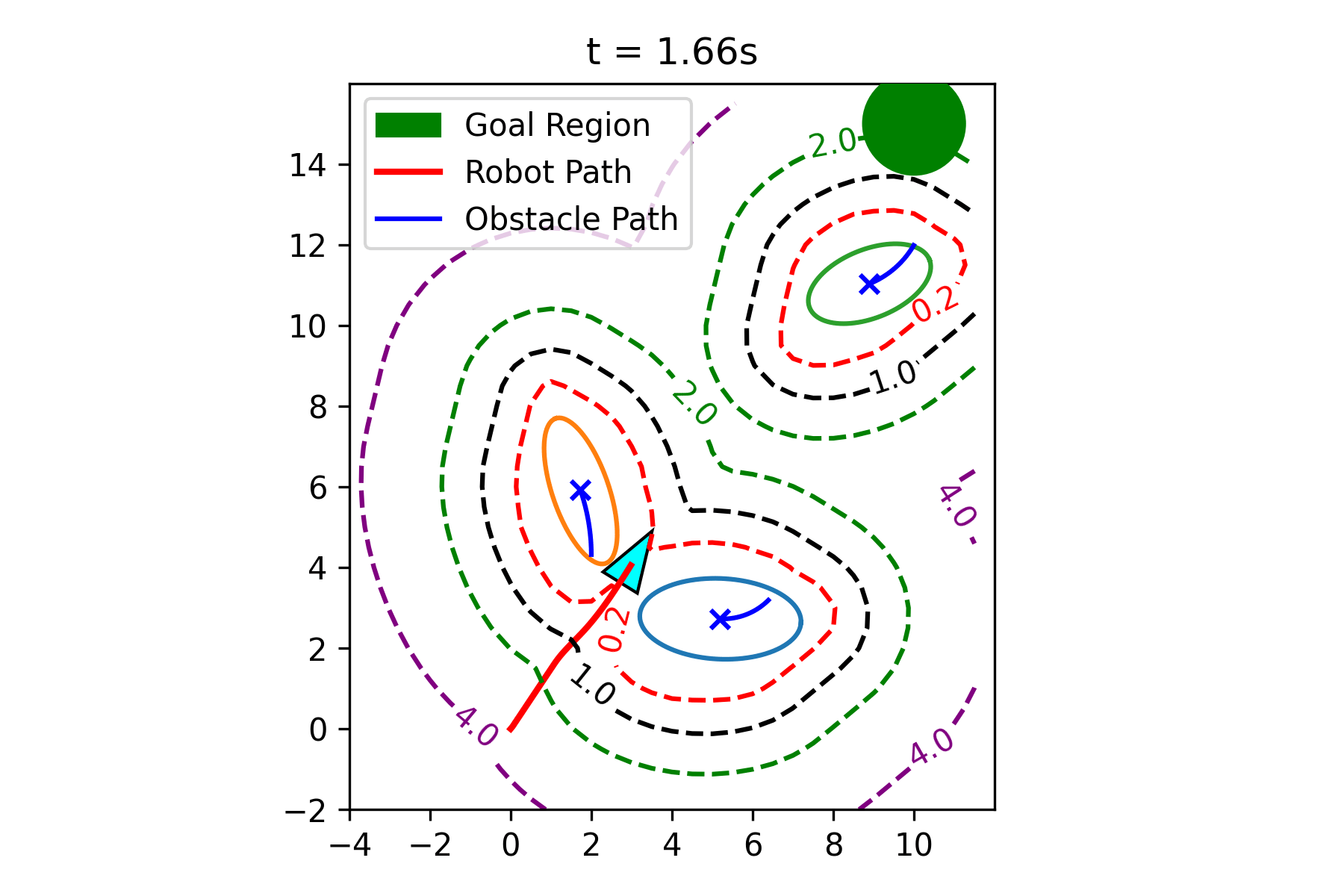}}
    \subcaptionbox{Time t = 4.12 sec \label{fig:2c}}
    {\includegraphics[width=0.19\textwidth, trim={3cm 0.2cm 3cm 0.0cm},clip]{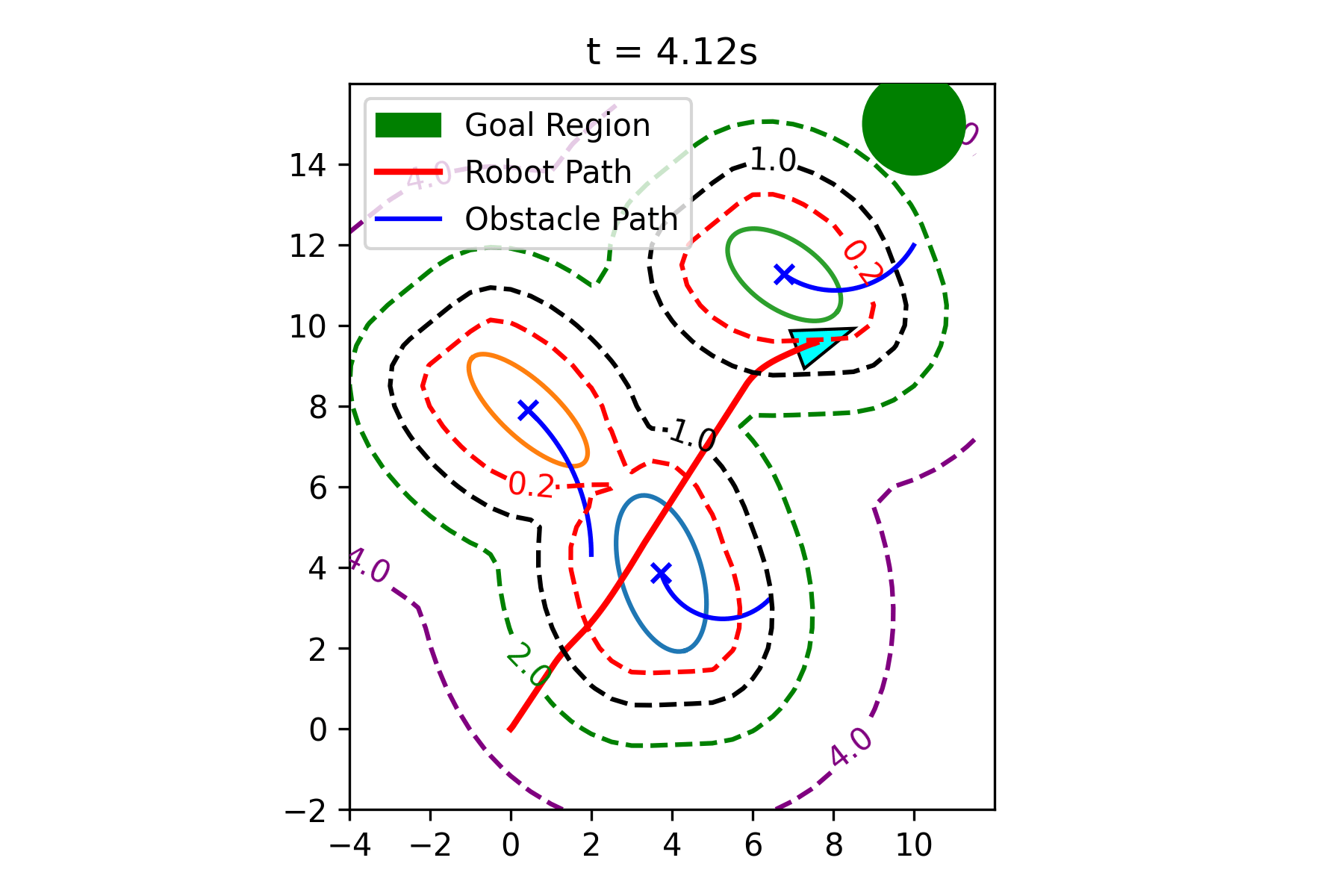}}
    \subcaptionbox{Final Pose \label{fig:2d}}
    {\includegraphics[width=0.19\textwidth, trim={3cm 0.2cm 3cm 0.0cm},clip]{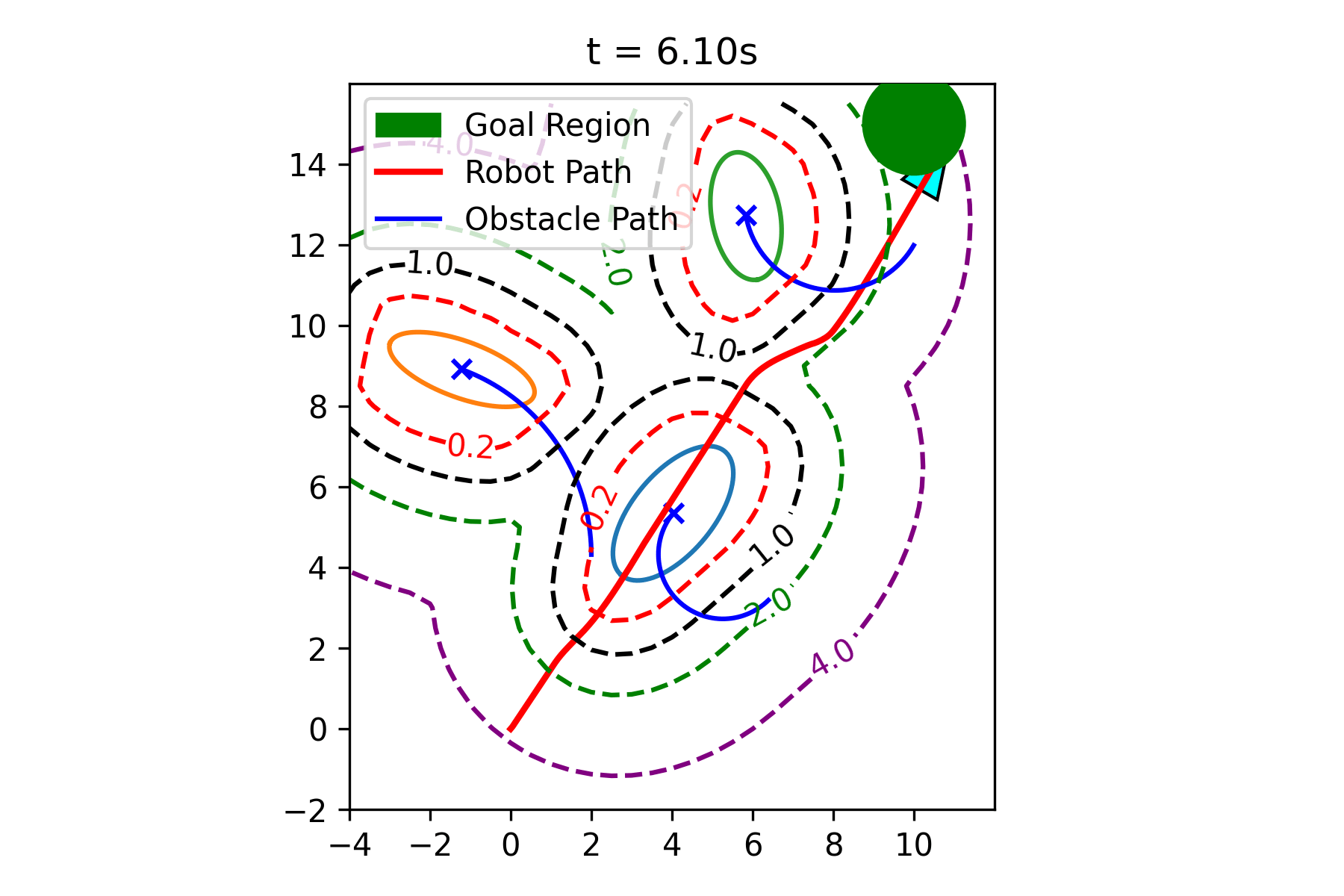}}
    \subcaptionbox{Circular Robot \label{fig:2e}}
    {\includegraphics[width=0.19\textwidth, trim={3cm 0.2cm 3cm 0.0cm},clip]{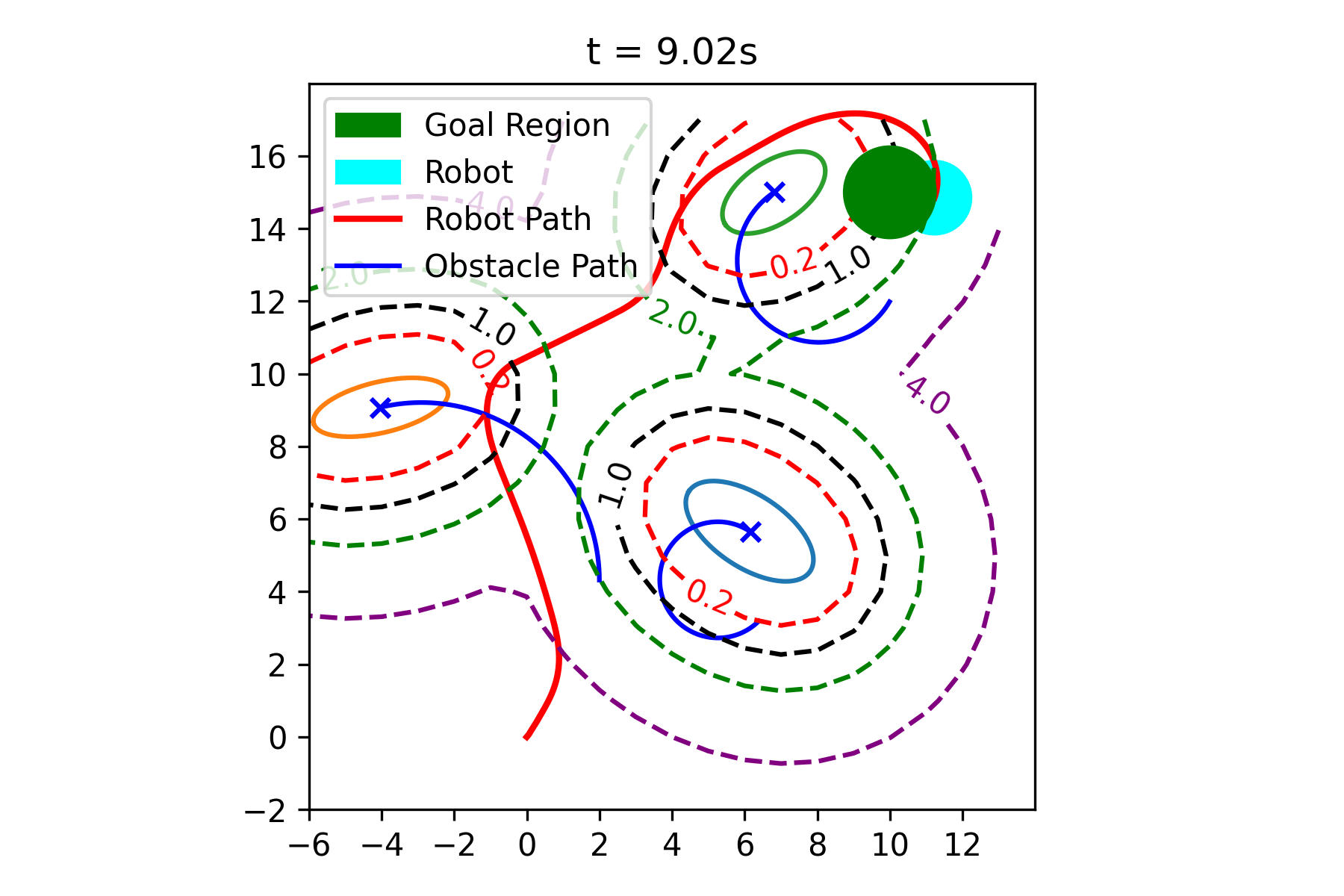}}
    
    % Space between rows
    %\vspace{-1mm}
    
    % Second row
    % \includegraphics[width=0.19\textwidth, trim={3cm 0.2cm 3cm 0.0cm},clip]{fig/unicycle/circular_snapshot_0.png}
    % \includegraphics[width=0.19\textwidth, trim={3cm 0.2cm 3cm 0.0cm},clip]{fig/unicycle/circular_snapshot_40.png}
    % \includegraphics[width=0.19\textwidth, trim={3cm 0.2cm 3cm 0.0cm},clip]{fig/unicycle/circular_snapshot_82.png}
    % \includegraphics[width=0.19\textwidth, trim={3cm 0.2cm 3cm 0.0cm},clip]{fig/unicycle/circular_snapshot_124.png}
    % \includegraphics[width=0.19\textwidth, trim={3cm 0.2cm 3cm 0.0cm},clip]{fig/unicycle/circular_snapshot_176.png}
    
    \caption{Safe navigation in a dynamical elliptical environment. (a) shows the initial pose of the triangular robot and the environment. (b) shows the triangular robot passing through the narrow space between two moving ellipses. (c) shows the robot adjusts its pose to avoid the moving obstacle. (d) shows the final pose of the robot that reaches the goal region. In (e), we plot the trajectory of navigating a circular robot in the same environment.}
    \label{fig:safe_navigation}
\end{figure*}

Suppose the robot has a polygonal shape with $\{\tilde{\bfp}_i\}$ denoting the vertices, and governed by unicycle kinematics,
\begin{equation}
\label{eq: unicycle_model}
\begin{bmatrix} \dot{x} \\ \dot{y} \\ \dot{\theta} \end{bmatrix} = \begin{bmatrix}\cos(\theta) &0 \\ \sin(\theta)  &0\\ 0 &1  \end{bmatrix} \begin{bmatrix}
    v \\ \omega
\end{bmatrix},
\end{equation}
where $v$, $\omega$ represent the robot linear and angular velocity, respectively. The state and input are $\bfx := [x , y ,\theta]^\top \in \mathbb{R}^2 \times [-\pi,\pi)$, $\bfu := [v,\omega]^\top \in \mathbb{R}^2$. The CLF for the unicycle model is defined as a quadratic form $
V(\bfx) = (\bfx - \bfx^*)^\top \mathbf{Q} (\bfx - \bfx^*)$, where $\bfx^*$ denotes the desired equilibrium and $\bfQ$ is a positive-definite matrix~\cite{unicycle_clf}. We define the goal region $\calG$ as a disk centered at the 2D position of the desired state $\bfx^*$, with a radius $r$.
 
By writing the robot's position as $ \tilde{\bfq} = [x,y]^\top $ and its orientation via the rotation matrix $ \tilde{\bfR}(\theta) $, we write the shape $S(\bfx)$ of the robot in terms of its state:
\begin{equation}
    S(\bfx) := \text{conv}\{\tilde{\bfq} + \tilde{\bfR}(\theta) \tilde{\bfp_i}\} 
\end{equation}
where $\tilde{\bfp_i}$ denotes the vertices of the polygon and $\text{conv}\{ \cdot \}$ denotes the convex hull of points. With this definition, we can derive the CBF for the polygon-shaped unicycle model, as in \eqref{eq: tv_cbf_define}.

\subsection{K-joint Robot Arm Safe Stabilizing Control}

\begin{figure*}[h]
    \centering
    % First row
    \subcaptionbox{Initial Pose \label{fig:3a}}
    {\includegraphics[width=0.19\textwidth, trim={0.5cm 0.2cm 2cm 0.0cm},clip]{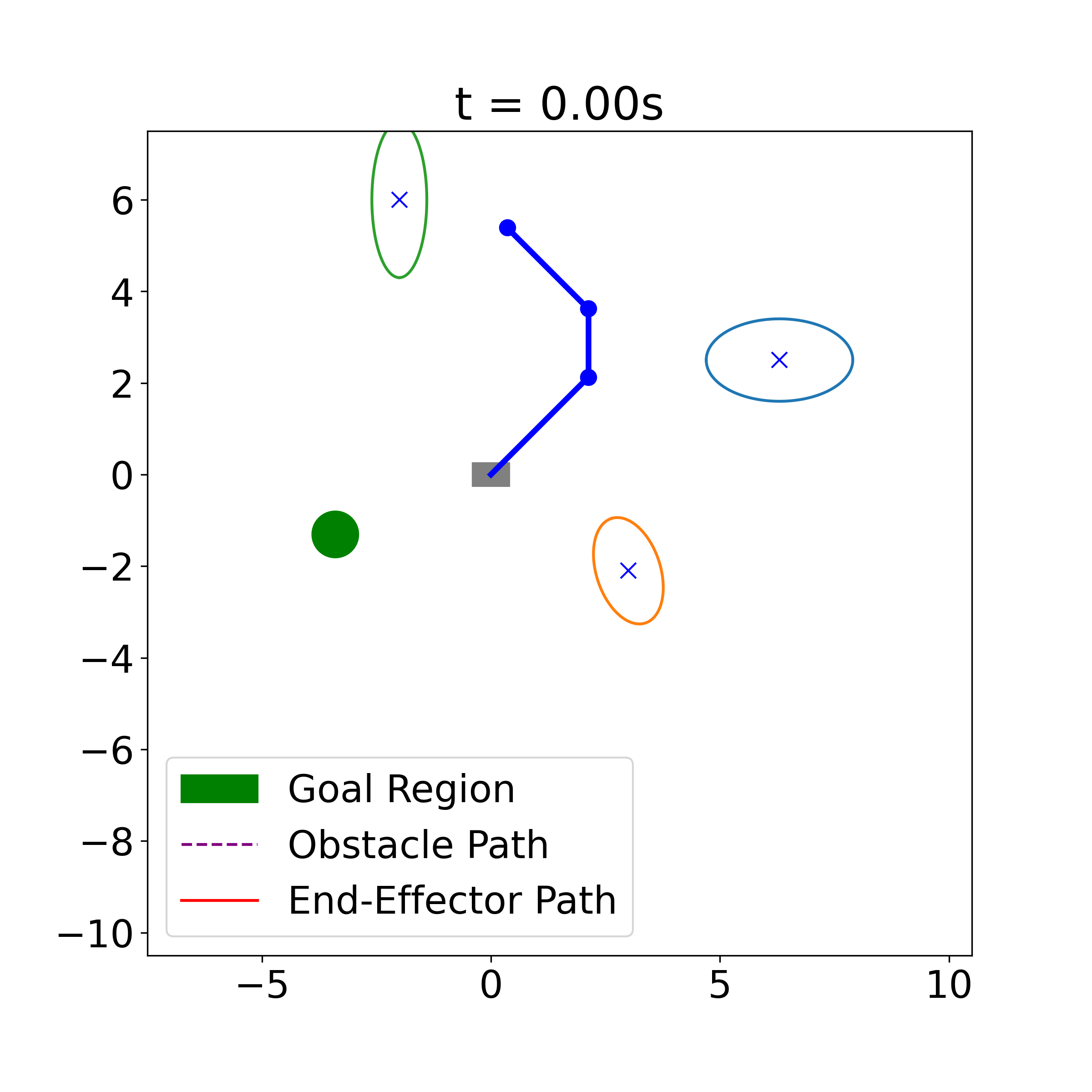}}
    \subcaptionbox{Time t = 4.12 sec \label{fig:3b}}
    {\includegraphics[width=0.19\textwidth, trim={0.5cm 0.2cm 2cm 0.0cm},clip]{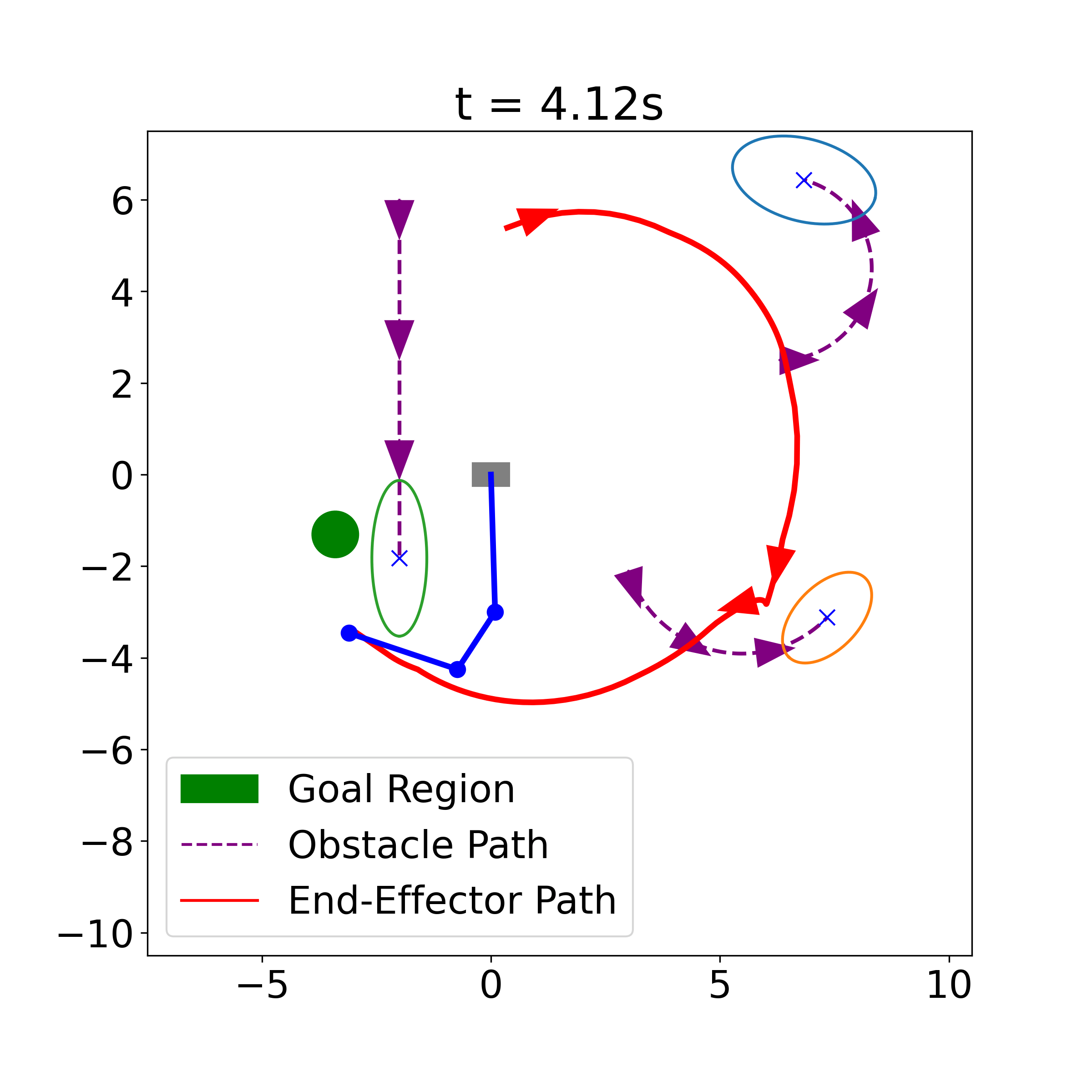}}
    \subcaptionbox{Time t = 4.92 sec \label{fig:3c}}
    {\includegraphics[width=0.19\textwidth, trim={0.5cm 0.2cm 2cm 0.0cm},clip]{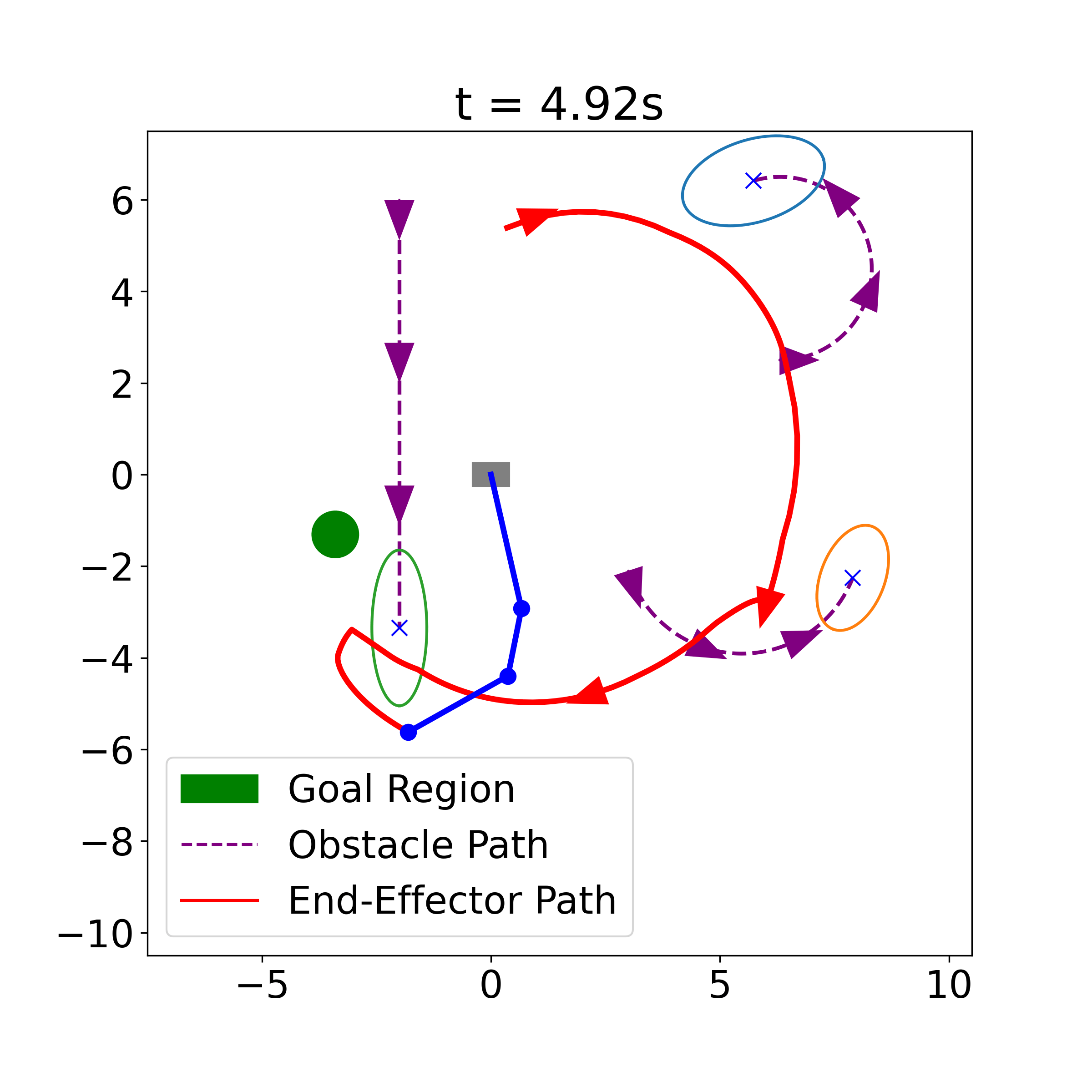}}
    \subcaptionbox{Time t = 6.28 sec  \label{fig:3d}}
    {\includegraphics[width=0.19\textwidth, trim={0.5cm 0.2cm 2cm 0.0cm},clip]{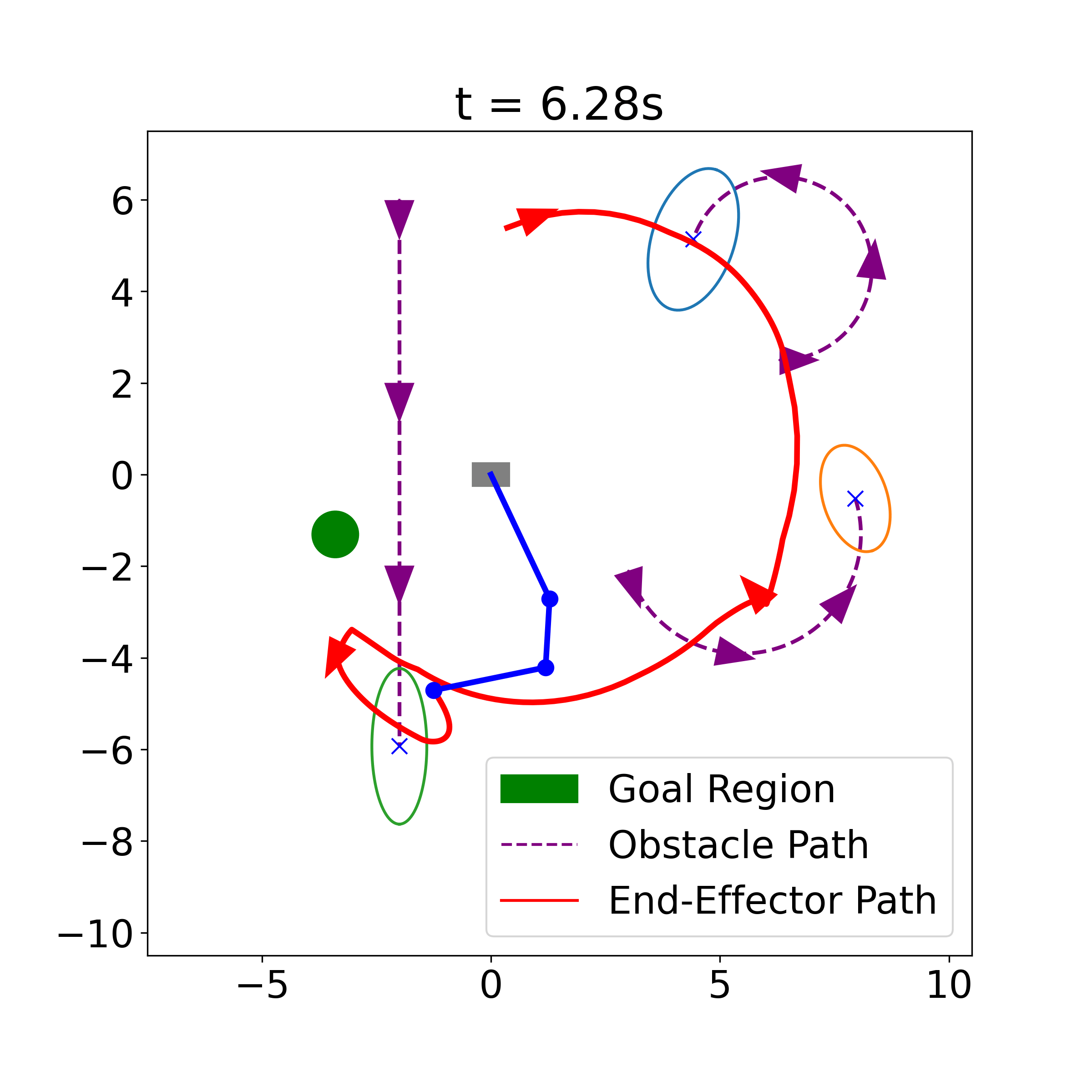}}
    \subcaptionbox{Final Pose \label{fig:3e}}
    {\includegraphics[width=0.19\textwidth, trim={0.5cm 0.2cm 2cm 0.0cm},clip]{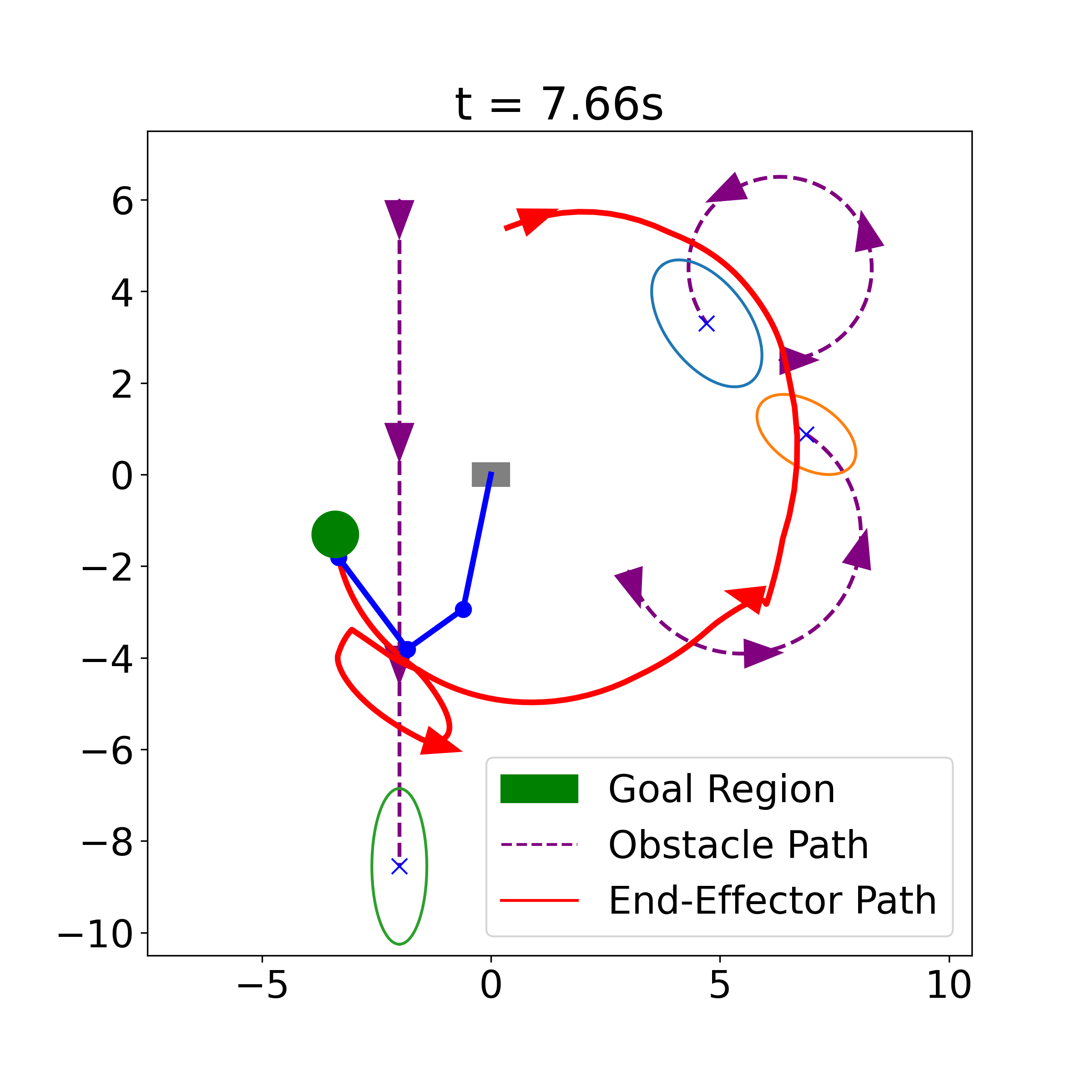}}
    \caption{Safe stabilization of a 3-joint robot arm. The green circle denotes the goal region, and the gray box denotes the base of the arm. The arm is shown in blue and the trajectory of its end-effector is shown in red. The trajectories of the moving elliptical obstacles are shown in purple.}
    \label{fig:robot_arm_safety}
    \vspace*{-3ex}
\end{figure*}

In this section, we discuss methods for controlling a 2D K-joint robot arm in a dynamical ellipse environment by utilizing our proposed CBF construction approach. For such robots, the links are intrinsically interconnected due to kinematic chaining. This means that controlling any one link will influence the pose of all subsequent links.

The dynamics of the robot arm are captured by:
\begin{equation}
\label{eq: M_d_arm_dynamics}
    \dot{\boldsymbol{\theta}} = \boldsymbol{\omega},
\end{equation}
where \( \boldsymbol{\theta} = [\tilde{\theta}_1, \tilde{\theta}_2, \ldots, \tilde{\theta}_K]^\top \) and \( \boldsymbol{\omega} = [\omega_1, \omega_2, \ldots, \omega_K]^\top \).

For the robot arm, each link has an associated 2D shape, denoted as $ S_i (\boldsymbol{\theta}) $, which depends on the state of the arm. The overall shape of the robot arm, is given by the union of these shapes $S(\boldsymbol{\theta}) = \bigcup_{i=1}^{K} S_i(\boldsymbol{\theta})$. For simplicity, we assume each $S_i $ is a line segment.

For each link \(i\), its state in $SE(2)$ consists of a position $\tilde{\bfq}_i = [x_i,y_i]^\top$:
%\begin{align}
$x_i = x_{i-1} + L_i \cos\left(\sum_{j=1}^{i} \tilde{\theta}_j\right)$ and
$y_i = y_{i-1} + L_i \sin\left(\sum_{j=1}^{i} \tilde{\theta}_j\right)$,
%\end{align}
%
and a rotation matrix $\tilde{\bfR}_i$ corresponding to $\underline{\theta_i} := \sum_{j=1}^{i}\tilde{\theta}_j$. For simplicity, we suppose $ x_1 = 0 $ and $ y_1 = 0 $, and $ L_i $ represents the length of the $i$-th link. The robot state can also be represented as multiple $ SE(2) $ configurations corresponding to each link, from $ (\tilde{\mathbf{q}}_1, \tilde{\bfR}_1) $ to $ (\tilde{\mathbf{q}}_{K}, \tilde{\bfR}_{K}) $. Additionally, we denote $\tilde{\mathbf{q}}_{K+1}$ as the end effector.

We define the CLF for the K-joint robot arm as 
$
V(\bftheta) = (\bftheta - \bftheta^*)^\top \mathbf{Q} (\bftheta - \bftheta^*),
$
where \( \mathbf{Q} \) is a positive-definite matrix, and \( \bftheta^*\) is the desired joint states. The goal region $\calG$ is specified as a disk centered at the position of the end effector corresponding to state $\bftheta^*$, with a defined radius $r$.

For safety assurance, the CBF is constructed using the distance between the robot arm and elliptical obstacles:
\begin{equation}
h_i(\bftheta) = \min_{j \in [K]} \Phi(\bfq_i, \bfR_i, \tilde{\mathbf{q}}_j, \tilde{\bfR}_j).
\end{equation}

% \NA{It is not clear how the orientation of the $i$-th link in the work space and its polygonal shape are related to the joint angles. This presentation makes it seem that the link orientation is the same as the joint angle, which is not true. I am missing some relationship between link shape, link pose, and joint angles, e.g., given by forward kinematics like Ch. 3.2 here \url{https://www.cse.lehigh.edu/~trink/Courses/RoboticsII/reading/murray-li-sastry-94-complete.pdf}.} \NA{What happens when there are singularities in the forward kinematics? It this CBF relative degree 1? See the discussion on manipulability in Ch. 5 of Lynch and Park.}
%

%% file: tex/Evaluation.tex
\section{Evaluation}
\label{sec: evaluate}

In this section, we show the efficacy of our proposed CBF construction techniques using simulation examples, focusing on ground-robot navigation and 2-D robot arm control.

Fig.~\ref{fig:se2_r2_compare} contrasts the $SE(2)$ distance function with the $\mathbb{R}^2$ counterpart by visualizing their level sets. Our proposed $SE(2)$ approach incorporates the orientation of the rigid-body robot, yielding notably improved results, particularly when the robot is close to obstacles.

To highlight the significance of accurate robot shape representation, we draw a comparison with a baseline circular robot CBF formulation. In Fig.~\ref{fig:safe_navigation}, we compare safe navigation using our proposed $SE(2)$ CBF approach with a regular $\mathbb{R}^2$ CBF approach. For both methods,  we set ${\bfk}(\bfx) = [v_{\max}, 0]^\top$ where $v_{\max} = 3.0$ is the maximum linear velocity. The remaining parameters were $\lambda = 100$, $\alpha_V(V(\bfx)) = 2V(\bfx)$, and $\alpha_h(h(\bfx, t)) = 3 h(\bfx, t)$.

We demonstrate safe navigation to a goal state. In Fig.~\ref{fig:2a}, the triangular robot starts the navigation with position centered at $(0,0)$ and orientation $\theta = \pi / 4$. In Fig.~\ref{fig:2b}, the robot adeptly navigates the narrow passage between two dynamic obstacles. In Fig.~\ref{fig:2d}, we see that the robot is able to reach the goal region without collision. In Fig.~\ref{fig:2e}, when the robot is conservatively modeled as a circle navigating the identical environment, it is evident that the robot has to opt for a more circuitous route to circumvent obstacles. This is due to its inability to traverse certain constricted spaces, as illustrated in Fig.~\ref{fig:2b}. These outcomes underscore the superior performance of our $SE(2)$ CBF methodology. Another advantage of the $SE(2)$ formulation lies in its assurance of a uniformly relative degree of $1$ for the constructed CBF, obviating the need to model a point off the wheel axis~\cite{cortes2017coordinated}.

% \begin{figure*}[h]
%     \centering
%     % First row
%     \subcaptionbox{Initial Pose \label{fig:3a}}
%     {\includegraphics[width=0.19\textwidth, trim={0.5cm 0.2cm 2cm 0.0cm},clip]{fig/robot_arm/arm_snapshot_0.png}}
%     \subcaptionbox{Time t = 4.12 sec \label{fig:3b}}
%     {\includegraphics[width=0.19\textwidth, trim={0.5cm 0.2cm 2cm 0.0cm},clip]{fig/robot_arm/arm_snapshot_206.png}}
%     \subcaptionbox{Time t = 4.92 sec \label{fig:3c}}
%     {\includegraphics[width=0.19\textwidth, trim={0.5cm 0.2cm 2cm 0.0cm},clip]{fig/robot_arm/arm_snapshot_246.png}}
%     \subcaptionbox{Time t = 6.28 sec  \label{fig:3d}}
%     {\includegraphics[width=0.19\textwidth, trim={0.5cm 0.2cm 2cm 0.0cm},clip]{fig/robot_arm/arm_snapshot_314.png}}
%     \subcaptionbox{Final Pose \label{fig:3e}}
%     {\includegraphics[width=0.19\textwidth, trim={0.5cm 0.2cm 2cm 0.0cm},clip]{fig/robot_arm/arm_snapshot_383.png}}
%     \caption{Safe stabilization of a 3-joint robot arm. The green circle denotes the goal region, and the gray box denotes the base of the arm. The arm is shown in blue and the trajectory of its end-effector is shown in red. The trajectories of the moving elliptical obstacles are shown in purple.}
%     \label{fig:robot_arm_safety}
% \end{figure*}

\begin{figure}
    \centering
    \includegraphics[width = 0.48\textwidth]{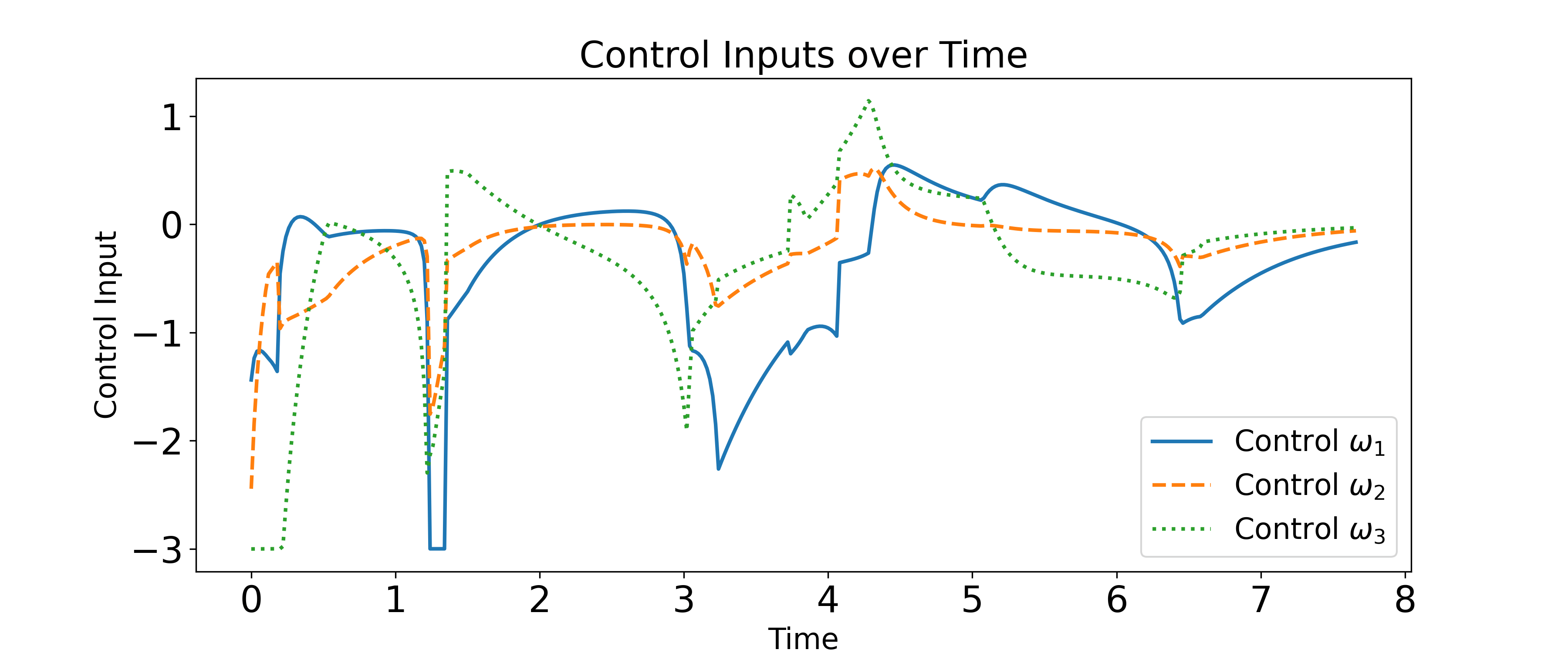}
    \caption{Control input of the 3-joint robot arm. }
    \label{fig:robot_arm_control_input}
    \vspace*{-3ex}
\end{figure}

\begin{figure}
    \centering
    \includegraphics[width = 0.48\textwidth]{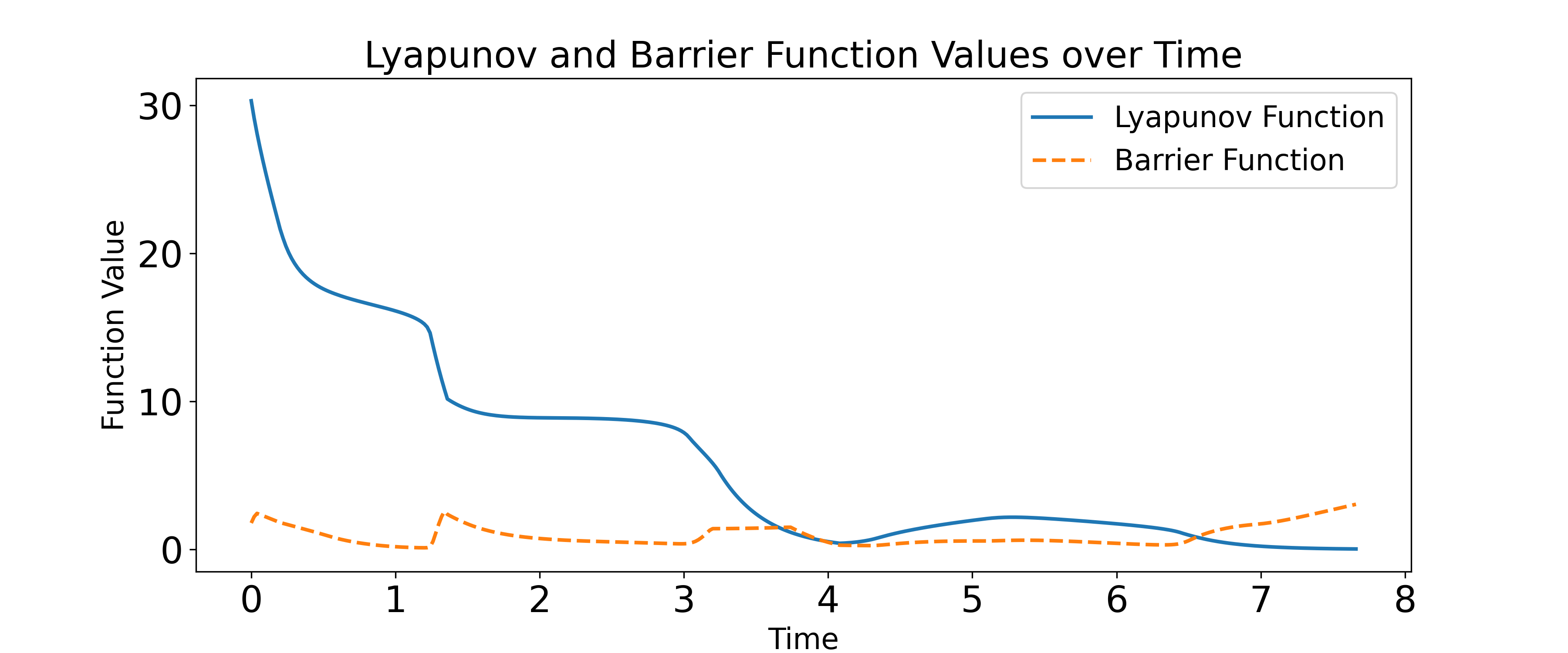}
    \caption{Lyapunov function and barrier function values over time.}
    \label{fig:robot_arm_LF_BF}
    \vspace*{-3ex}
\end{figure}

In the following set of experiments, we consider safe stabilization of a 3-joint robot arm in a dynamical elliptical environment. We set ${\bfk}(\bfx) = [0, 0, 0]^\top$ and restrict the joint control bounds with $|\omega_i | \leq 3$. In Fig.~\ref{fig:robot_arm_safety}, the robot arm is able to elude the mobile ellipses by nimbly adjusting its pose. In Fig.~\ref{fig:robot_arm_control_input}, we show the control inputs of each joint over time. We see that when the robot arm is close to the obstacles, it is able to take large control inputs in adjusting its pose. In Fig.~\ref{fig:robot_arm_LF_BF}, we show the CLF and CBF values over time. A consistently positive CBF value throughout the trajectory signifies safety assurance, while the decreasing CLF values indicates the convergence to the desired state. Moreover, the CLF value may increase when the arm is close to obstacles (i.e. CBF value is low), this comes from the relaxation of the CLF-CBF QP to ensure the feasibility of the program.

%% file: tex/Conclusion.tex
\section{Conclusion}

We present an analytic distance formula between elliptical and polygonal objects.  Leveraging this formula, we construct a time-varying control barrier function that ensures the safe autonomy of a polygon-shaped robot operating in dynamical elliptical environments. The efficacy of the proposed approach is demonstrated in rigid-body navigation and multi-link robot arm problems. 
% \& Future Work
Future work will consider extending the formulation to 3-D robot arm manipulation and estimating the geometry and dynamics of the environment with on-board sensing.